\documentclass[11pt]{article}

\usepackage{fullpage}
\usepackage[ruled, linesnumbered, vlined, commentsnumbered]{algorithm2e}
\usepackage{graphicx,subfig} 
\usepackage{amsfonts,amssymb,amsmath,amsthm,amsopn,mathtools}	
\usepackage{booktabs,diagbox,colortbl,multirow,tabularx,threeparttable,hhline}
\usepackage[listings,skins,breakable]{tcolorbox}

\usepackage{enumerate}
\usepackage[shortlabels]{enumitem}
\usepackage{csquotes}
\usepackage{authblk}
\usepackage{footnote}
\usepackage{hyperref}
\usepackage{prettyref}
\usepackage{cite}
\usepackage{setspace}
\usepackage{color}
\usepackage{xcolor}  
\usepackage{scrpage2}
\usepackage{geometry}

\usepackage{tikz}
\usepackage{pgfplots}
\usetikzlibrary{positioning,shapes,shadows,arrows,calc}
\tikzstyle{component}=[rectangle, draw=black, rounded corners, fill=blue!40, drop shadow, text centered, anchor=north, text=white, minimum height=1cm]
\tikzstyle{arrow}=[->, thick]

\pgfplotsset{compat=1.12}
\usetikzlibrary{intersections}
\usetikzlibrary{pgfplots.statistics}
\usepgfplotslibrary{fillbetween}

\definecolor{myblue}{RGB}{34,31,217}
\definecolor{mycyan}{gray}{.7}
\definecolor{Gray}{gray}{0.9}

\newtheorem{remark}{Remark}

\newtcolorbox{quotebox}{colback=gray!10,boxrule=0.4pt,colframe=black,fonttitle=\bfseries,top=1pt,bottom=1pt}

\def\our{\texttt{\textsc{DaNuoYi}}}

\SetCommentSty{mycommfont}

\newcommand{\bb}[1]{\multicolumn{1}{>{\columncolor{mycyan}}c}{\textbf{{#1}}}}

\newcommand{\pref}{\prettyref}

\newrefformat{fig}{Fig.~\ref{#1}}
\newrefformat{tab}{Table~\ref{#1}}
\newrefformat{sec}{Section~\ref{#1}}
\newrefformat{alg}{Algorithm~\ref{#1}}
\newrefformat{property}{Property~\ref{#1}}
\newrefformat{theorem}{Theorem~\ref{#1}}
\newrefformat{definition}{Definition~\ref{#1}}
\newrefformat{corollary}{Corollary~\ref{#1}}
\newrefformat{lemma}{Lemma~\ref{#1}}
\newrefformat{conj}{Conjecture~\ref{#1}}
\newrefformat{def}{Definition~\ref{#1}}
\newrefformat{eq}{equation~(\ref{#1})}
\newrefformat{app}{Appendix~\ref{#1}}

\usepackage{lscape}

\newenvironment{code-example}
{
\vspace{0.15cm}
\noindent\begin{minipage}{\linewidth}
\begin{center}
\arrayrulecolor{black}
\color{black}
\begin{tabular}{|p{0.95\linewidth}|}
\hline%
\rowcolor{pink!20}%
}
{
\\\hline
\end{tabular}
\end{center}
\end{minipage}
\vspace{-0.2cm}
}

\begin{document}

\title{\vspace{-1ex}\LARGE\textbf{\our: Evolutionary Multi-Task Injection Testing on Web Application Firewalls}~\footnote{This manuscript is submitted for potential publication. Reviewers can use this version in peer review.}}

\author[1]{\normalsize Ke Li}
\author[1]{\normalsize Heng Yang}
\author[2]{\normalsize Willem Visser}
\affil[1]{\normalsize Department of Computer Science, University of Exeter, EX4 4QF, Exeter, UK}
\affil[2]{\normalsize Amazon Web Service, USA}
\affil[$\ast$]{\normalsize Email: \texttt{k.li@exeter.ac.uk}}

\date{}
\maketitle

\vspace{-3ex}
{\normalsize\textbf{Abstract: } }Web application firewall (WAF) plays an integral role nowadays to protect web applications from various malicious injection attacks such as SQL injection, XML injection, and PHP injection, to name a few. However, given the evolving sophistication of injection attacks and the increasing complexity of tuning a WAF, it is challenging to ensure that the WAF is free of injection vulnerabilities such that it will block all malicious injection attacks without wrongly affecting the legitimate message. Automatically testing the WAF is, therefore, a timely and essential task. In this paper, we propose \textsc{DaNuoYi}, an automatic injection testing tool that simultaneously generates test inputs for multiple types of injection attacks on a WAF. Our basic idea derives from the cross-lingual translation in the natural language processing domain. In particular, test inputs for different types of injection attacks are syntactically different but may be semantically similar. Sharing semantic knowledge across multiple programming languages can thus stimulate the generation of more sophisticated test inputs and discovering injection vulnerabilities of the WAF that are otherwise difficult to find. To this end, in \textsc{DaNuoYi}, we train several injection translation models by using multi-task learning that translates the test inputs between any pair of injection attacks. The model is then used by a novel multi-task evolutionary algorithm to co-evolve test inputs for different types of injection attacks facilitated by a shared mating pool and domain-specific mutation operators at each generation. We conduct experiments on three real-world open-source WAFs and six types of injection attacks, the results reveal that \textsc{DaNuoYi} generates up to $3.8\times$ and $5.78\times$ more valid test inputs (i.e., bypassing the underlying WAF) than its state-of-the-art single-task counterparts and the context-free grammar-based injection construction.



\section{Introduction}
\label{sec:introduction}

Due to the maturity of the state-of-the-art web technologies and advancements of Internet of things, web applications have become increasingly ubiquitous and important for enterprises and individuals from various sectors, such as online shopping, e-banking, healthcare, e-governance and social media. Yet, the prevalence of web applications inevitably make them one of the main targets of malicious attacks. For example, Beery and Niv~\cite{beery2013web} reported that each web application worldwide can experience around $173$ injection attacks per month on average. According to a recent report published by Open Web Application Security Project (OWASP)\footnote{\url{https://owasp.org/Top10/}}, injection attack is one of the most common way to compromise a web application and its data.

An unified and state-of-the-practice solution to injection attacks is the use of a web application firewall (WAF)~\cite{AppeltNPB18}, which is a special type of application firewall that has been widely adopted to provide protection of web applications from various malicious injection attacks. It is usually deployed as a facade between the client and web application server aiming to analyze all HTTP messages, which contain potentially malicious user inputs, sent to the web application --- detecting, filtering and blocking anything malicious through a set of rules. As such, a WAF is application-independent and is designed to prevent any type of injection attack in mind. By a type of injection attack, we refer to the attack that specifically seeks to inject malicious code into a particular programming language used in a web application, such as SQL injection (SQLi), XML injection (XMLi), and PHP injection (PHPi).

Given the fast-moving nature of web applications, injection attacks relentlessly emerge all the time and are grown with an evolving sophistication. As a result, fine-tuning the rules in a WAF is a complex, labor-intensive and costly task especially in the presence of multiple types of injection attack. Note that a reliable WAF not only needs to be able to detect the increasingly sophisticated injection attacks but also to avoid mistakenly blocking legitimate HTTP messages. As such, this leaves the WAF with a great chance to suffer from various injection vulnerabilities in practice.

\begin{table*}[t!]
\caption{Examples of three semantically similar but syntactically different types of injection attacks}
\label{tab:exp}
\centering
\footnotesize
\begin{tabular}{ccl}\toprule
\multirow{2}{*}{SQLi}&vulnerable code&\texttt{SELECT * FROM users WHERE username='\textcolor{blue!50}{\textbf{\$name}}' and password='\textcolor{blue!50}{\textbf{\$pass}}';}\\
&injection attack&\texttt{SELECT * FROM users WHERE username='\textcolor{blue!50}{\textbf{' OR '1'='1';}}\colorbox{pink}{\textcolor{blue!50}{\textbf{ --}}' and password='\textcolor{blue!50}{\textbf{abc}}';}}\\\hline

\multirow{2}{*}{XMLi (XPathi)}&vulnerable code&\texttt{users[username/text()='\textcolor{blue!50}{\textbf{\$name}}' and password/text()='\textcolor{blue!50}{\textbf{\$pass}}']}\\
&injection attack&\texttt{users[username/text()='\textcolor{blue!50}{\textbf{' OR '1'='1}}\colorbox{pink}{\textcolor{blue!50}{\textbf{(:}}' and password/text()='\textcolor{blue!50}{\textbf{:)}}}']}\\\hline


\multirow{2}{*}{PHPi}&vulnerable code&\texttt{if (\textit{eval}("return '\$storedName' === '\textcolor{blue!50}{\textbf{\$name}}' \&\& '\$storedPass' === '\textcolor{blue!50}{\textbf{\$pass}}';"))}\\

&injection attack&\texttt{return \$storedName === '\textcolor{blue!50}{\textbf{' || '1'=='1';}} \colorbox{pink}{\textcolor{blue!50}{\textbf{//}}' \&\& \$storedPass === '\textcolor{blue!50}{\textbf{abc}}';}}\\

\bottomrule
\end{tabular}
\centering
 \begin{tablenotes}
    \footnotesize
     \item The \textcolor{blue!50}{blue}/bold texts are the inputs from the user, where \textcolor{blue!50}{\textbf{\$name}} and \textcolor{blue!50}{\textbf{\$pass}} denote the variable of the username and password given by an user, respectively. The highlighted texts are what will be commented out by the injection attacks.
    \end{tablenotes}
\end{table*}

To verify whether a WAF is tuned to be sufficiently secured before its production deployment, a variety of testing techniques has been proposed for generating test inputs to the WAF, including white-box testing~\cite{GodefroidKL08}, static analysis~\cite{FuLPCQT07}, model-based testing~\cite{JurjensW01} and black-box testing~\cite{DoupeCV10}. However, none of these techniques are perfect because they are either less applicable in practice or inadequate for vulnerability detection. For example, both white-box testing methods and static analysis tools require full control of source codes, the access of which is difficult, if not impossible, in web applications and their WAFs due to the heterogeneous programming environments~\cite{ShinMWO11}. Henceforth, it is not difficult to understand that the detection capability of white-box testing is limited. As for the model-based testing techniques, neither developing models expressing the security policies nor constructing the implementation of WAFs and the web applications is easily accessible. Although black-box testing tools~\cite{AppeltPB17,AppeltNPB18,JanPAB19}, mostly fuzz testing, do not require the access of source code, they often focus on the syntax of attacks yet ignoring the semantic information, which could restrict their testing capability.

A major bottleneck of the existing black-box testing tools for WAF is the lack of support for testing more than one injection attack simultaneously. This essentially contradicts the design principle of a WAF, which aims to serve as a universal filter that works independently of the programming language(s) that underpins a web application. Beside the impact on the applicability of the tools, such a limitation also shuts the gate towards achieving more effective injection testing. This is because, similar to human natural languages, malicious injection attacks of different programming languages may impose different syntax but do have some unique semantics in common. As such, analogous to cross-lingual translation, they can share certain common ground between them, which could generate more sophisticated test inputs to discover injection vulnerabilities that are otherwise difficult to find. Let us consider the examples shown in \pref{tab:exp} which gives three types of injection attack, i.e., SQLi, XMLi (XPath injection) and PHPi. Clearly, all these injection attacks contain malicious and syntactically different inputs, i.e., the \textcolor{blue}{\texttt{'1'='1'}}, \textcolor{blue}{\texttt{OR}}, and \textcolor{blue}{\texttt{--}} for SQLi; the \textcolor{blue}{\texttt{'1'='1'}}, \textcolor{blue}{\texttt{OR}}, and \textcolor{blue}{\texttt{(::)}} for XMLi; and the \textcolor{blue}{\texttt{'1'=='1'}}, \textcolor{blue}{\texttt{||}}, and \textcolor{blue}{\texttt{//}} for PHPi. However, the three injection attacks are also semantically similar in the sense that they all aim to \lq fool\rq\ the WAF and the web application by creating a tautology (i.e., the string \textcolor{blue}{\texttt{'1'}} is always the same) and commenting out parts of the original command fragments. This constitutes the key motivation behind this work.


To overcome the aforementioned limitations, this paper proposes an evolutionary multi-task injection testing fuzzer for WAF, dubbed \texttt{\textsc{DaNuoYi}}\footnote{\texttt{\textsc{DaNuoYi}} is originally from \textit{Heavenly Sword and Dragon Slaying Sabre}, a famous knights-errant novel by Jin Yong (Louis Cha). Its full name is \texttt{Qiankun Danuoyi} and it one of the most premium Kongfu which is able to leverage and mimic different types of Kongfu power from the enemies to attack them back harder --- this is similar to what our approach can do in imitating different types of injection attack to test the WAF.}, which automatically and simultaneously tests multiple types of injection attacks. Specifically, \texttt{\textsc{DaNuoYi}} uses the language models (i.e., Word2Vec, see \pref{sec:word_embedding}) to capture the semantic information of each type of injection attack. In particular, such semantic information is shared across different injection types during the training process thus to facilitate the translation of a test input from one type (e.g., SQLi) into another (e.g., PHPi) in a multi-task learning manner. Note that the learned translation models thereafter serve as the \lq bridges\rq\ between different injection testing tasks during the test input generation driven by a novel multi-task evolutionary algorithm. By doing so, \texttt{\textsc{DaNuoYi}} is empowered to improve the test input generation of one injection type by borrowing the promising test inputs generated for all other types considered.

\noindent\textbf{\underline{\textit{Contributions}}.}
In a nutshell, our contributions include:
\begin{itemize}
    \item \texttt{\textsc{DaNuoYi}} is a fully automatic, end-to-end tool that can simultaneously test any type of injection attack on a variety of WAFs. To the best of our knowledge, this is the first tool of its kind that can automatically and simultaneously generate test inputs from multiple types of injection attacks for testing a WAF.

    \item Surrogate classifiers based on neural networks are developed to embed the semantic information of injection contexts into a vector representation to predict the likelihood of bypassing the underlying WAF.

    \item A multi-task translation (i.e., multi-task injection translation) framework trained by an encoder-decoder architecture from the natural language processing (NLP) domain. Based on this framework, a test input for one type of injection attack can be translated into a syntactically different, but semantically similar one for another type.
    
    \item A multi-task evolutionary algorithm (MTEA), empowered by the surrogate classifiers, the multi-task translation module, and six tailored mutation operators, co-evolves test inputs for different types of injection attacks.
    
    \item A quantitative and qualitative analysis\footnote{The source code and data related to this paper can be found from our lab repository: \href{https://github.com/COLA-Laboratory/DaNuoYi}{https://github.com/COLA-Laboratory/DaNuoYi}} on three real-world WAFs (i.e., \texttt{ModSecurity}, \texttt{Ngx-Lua-WAF}, \texttt{Lua-Resty-WAF}, see \ref{subject_waf} for details), 
    three alternative classifiers, and six types of injection attacks including SQLi, XMLi, PHPi, HTML injection (HTMLi), OS shell script injection (OSi), cross-site script injection (XSSi). Empirical results fully demonstrate the superiority of \our\ for disclosing more injection vulnerabilities for WAF (up to $3.8\times$ more bypassing injection cases) comparing to the single-task counterparts.
\end{itemize}

\noindent\textbf{\underline{\textit{Novelty}}.} What makes \texttt{\textsc{DaNuoYi}} \textit{unique} are:

\begin{itemize}
    \item It is able to learn the common semantic information from syntactically different test inputs for different types of injection attacks. This is achieved by the translation model that mimics the cross-lingual translation between natural languages.
    
    \item It is capable of generating test inputs for any type of injection attacks with different syntax and exploit the most promising test inputs interchangeably. This is realized by a MTEA that co-evolves multiple test input generation processes simultaneously.
\end{itemize}

The rest of this paper is organized as follows. \pref{sec:background} provides some background knowledge related to the injection testing on WAF and the neural language model for injection testing. \pref{sec:approach} delineates the technical details of \our\ while its effectiveness is quantitatively and qualitatively analyzed on three open-source real-world WAFs as in~\pref{sec:evaluation}. At the end, Sections~\ref{sec:tov} and \ref{sec:related} discuss the threats to validity and related work respectively while \pref{sec:conclusion} concludes this paper and sheds some lights on future directions.


\section{Preliminaries}
\label{sec:background}

\subsection{Injection Testing on WAF}

In the development of modern web applications, a WAF serves as the first entity to filter malicious injection attacks coming to the underlying application. Taking the most commonly used signature-based WAFs as an example, this is achieved by setting some rules, e.g., the regular expression, that detects critical words in the user inputs embedded in a HTTP request it receives.

\begin{figure}[t!]
	\centering
	\includegraphics[width=0.6\columnwidth]{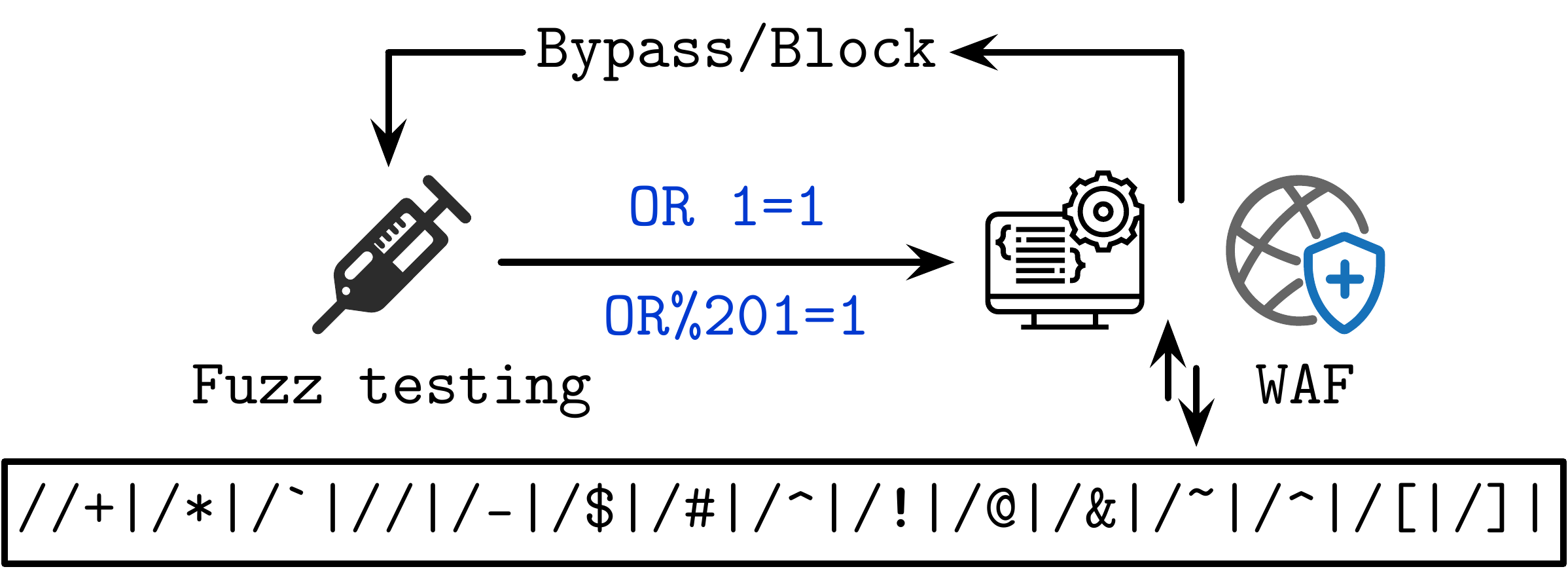}
	\caption{Illustrative example of injection testing for WAF.}
	\label{fig:sec2}
\end{figure}

\pref{fig:sec2} gives an intuitive example of the a typical testing process on a WAF. To test a particular type of injection, say SQLi, one usually relies on fuzzing that randomly generates test inputs for the WAF. In practice, a test input is a failed attack in case it is blocked by the WAF; otherwise it bypasses the WAF thus indicating a SQLi vulnerability under the current WAF setting. For example, the expression shown in~\pref{fig:sec2} can block any input that contains \textcolor{blue}{\texttt{+*`/-\$\#\^}}\textcolor{blue}{\texttt{!@\&$\thicksim$}}. Therefore, the test input \textcolor{blue}{\texttt{OR 1=1}} will be blocked. However, some other ones, e.g., \textcolor{blue}{\texttt{OR\%201=1}}, can bypass the WAF. Note that, according to Demetrio et al.~\cite{DemetrioVCL20} and Appelt et al.~\cite{AppeltNPB18}, the test inputs that bypass the WAF are always considered to be malicious, as they can either successfully inject the web application behind or would provide necessary information that help to eventually achieve so (e.g., allowing the attackers to know which inputs have been blocked or not --- a typical blind attack).

Although a HTTP message may contain multiple user inputs (e.g., a login form needs both username and password), the entire message is compromised as long as one of the malicious user inputs can bypass the WAF. Therefore, automatically testing WAF often focus on generating a single test input for the WAF~\cite{AppeltNPB18}. Further, it can be easily adopted to other cases where multiple user inputs are required by combining different generated test inputs together.

\subsection{Feature Embedding for Injection Testing}
\label{sec:bg-nlm}

A successful test input that discovers injection vulnerabilities needs to comply with the syntax of the underlying injection type. Our recent study demonstrated that a programming language bears many similarities with natural languages. Henceforth, leveraging the semantic information embedded in injection attacks can facilitate the automated test case generation accordingly.

To exploit the semantic information of a test input, the first step is to embed its words into a measurable representation. Its basic idea is to encode each word separated by blank, through a vocabulary, into a fixed length vector with the same dimension as the number of words in the test input. Taking SQLi as an example, the test input \textcolor{blue}{\texttt{' OR '1'='1'}} will be tokenized into \{\textcolor{blue}{\texttt{'}}, \textcolor{blue}{\texttt{OR}}, \textcolor{blue}{\texttt{'}}, \textcolor{blue}{\texttt{1}}, \textcolor{blue}{\texttt{'}}, \textcolor{blue}{\texttt{=}}, \textcolor{blue}{\texttt{'}}, \textcolor{blue}{\texttt{1}},\textcolor{blue}{\texttt{'}}\}. After distinguished 
encoding, \textcolor{blue}{\texttt{'}$=(1,0,0,0)$}, \textcolor{blue}{\texttt{OR}$=(0,1,0,0)$}, \textcolor{blue}{$\texttt{1}=(0,0,1,0)$}, and \textcolor{blue}{$==(0,0,0,1)$}. These vectors are then further embedded by a neural network, namely a neural language model (NLM), into another $d\geq 1$ dimensional vector ($d$ is a predefined hyper-parameter) to take additional contextual information into account (see Section~\ref{sec:word_embedding}). In particular, each dimension partially contributes to the meaning of a word. For example, when $d=5$, we may have \textcolor{blue}{\texttt{OR}$=(32,19,3,81,22)$}. The word vectors would then be further extracted, when needed, to better understand their semantic information in the translation, for which we will elaborate in Section~\ref{sec:seq2seq}.



\section{Multi-task Injection Testing with \our}
\label{sec:approach}

\begin{figure*}[t!]
	\centering
	\includegraphics[width=.8\textwidth]{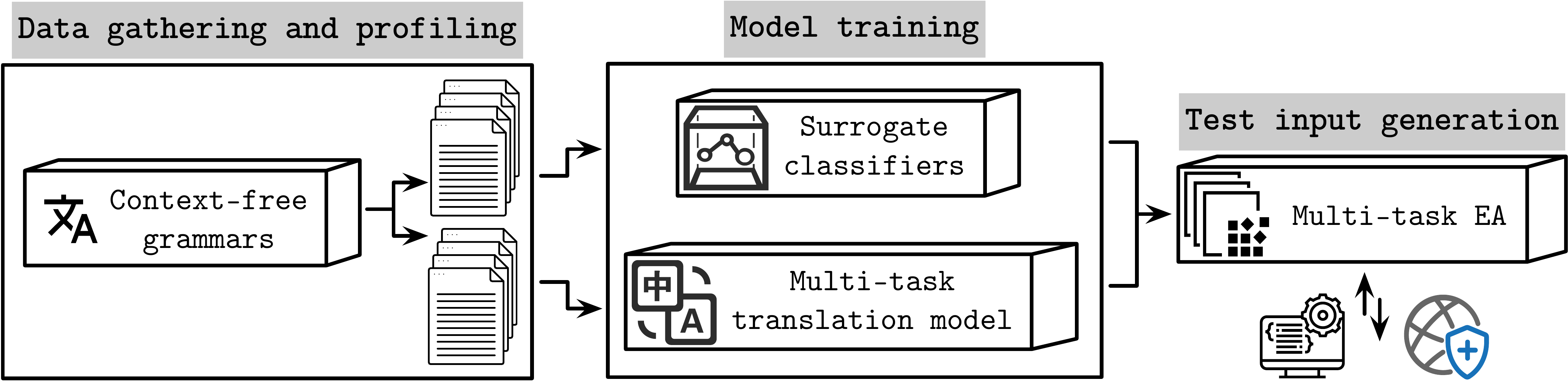}
	\caption{System architecture and workflow of \our.}
	\label{fig:DaNuoYi}
\end{figure*}

\our\ is designed as an end-to-end fuzzing tool to automate the test input generation for detecting multiple types of injection vulnerabilities. Its overarching hypothesis is that test inputs from syntactically different types of injection attacks share certain latent semantic similarities that can be useful to generate sophisticated test inputs for each other. As shown in~\pref{fig:DaNuoYi}, the main workflow in \our\ consists of four components, each of which is outlined as follows.

\begin{itemize}
   \item \textit{\underline{Data gathering and profiling:}} \our\ assumes a context-free grammar (CFG) for each type of injection attack, based on which test inputs can be generated to profile the WAF.
      
   \item \textit{\underline{Surrogate classifiers:}} To better distinguish the \lq good\rq\ test inputs from the \lq bad\rq\ ones for a WAF, in \our\, we build a classifier for each type of injection attack, by using the data generated by the CFG, to estimate the likelihood of a test input bypassing the WAF. Note that this likelihood is used to evaluate the fitness of a candidate test input generated by the MTEA.
   
   \item \textit{\underline{Multi-task translation model:}} We develop a multi-task injection translation paradigm to bridge the test input generation across different types of injection attack. For any pair of injection types, we build a translation model that translates the test input from one type of injection attack into a semantically related one. Note that all these translation models are trained using the data also generated by the CFG and their parameters are shared during the training process, thus we can expect to improve the effectiveness of the translated test inputs.
    
   \item \textit{\underline{Multi-task evolutionary algorithm:}} To generate test inputs for multiple types of injection attack simultaneously, we develop a MTEA to continuously evolve test inputs towards successful injections. The initial test inputs are seeded according to the CFG. During each evolutionary iteration, promising test inputs generated for one type of injection attack are shared across all other types of injection attack by using the multi-task translation models. Only the most promising test inputs, which are highly likely to achieve successful injections, can survive to the next iteration.
\end{itemize}

In the following paragraphs, we elaborate the implementation detail of each components.

\subsection{Data Gathering and Profiling}
\label{sec:data}

\subsubsection{Context-free Grammar for Injection Searching}
\label{sec:grammar}

Data is arguably the \textit{fountain of life} for any machine learning task. In \our, we take advantage of the CFG for each type of injection attack to automatically generate some initial test inputs, called rule-based injection search method. Note that for the types of injection attack considered in this work, the grammar for SQLi is derived from those defined in~\cite{AppeltNPB18} while the others are developed by ourselves according to a systematic summary of the literature\cite{SalasGM15,SuW06,HydaraSZA15} and several well known open-source payload\footnote{~\href{https://github.com/SpiderLabs/MCIR}{MCIR}, \href{https://github.com/xsscx/Commodity-Injection-Signatures}{Commodity-Injection-Signatures}, \href{https://github.com/payloadbox/}{xxe-injection-payload-list} }. In practice, the CFG for any arbitrary type of injection attack can be created in a similar way or there are readily available ones to use, such as those generated in this work.

\begin{figure*}[t!]
	\centering	\includegraphics[width=.8\linewidth]{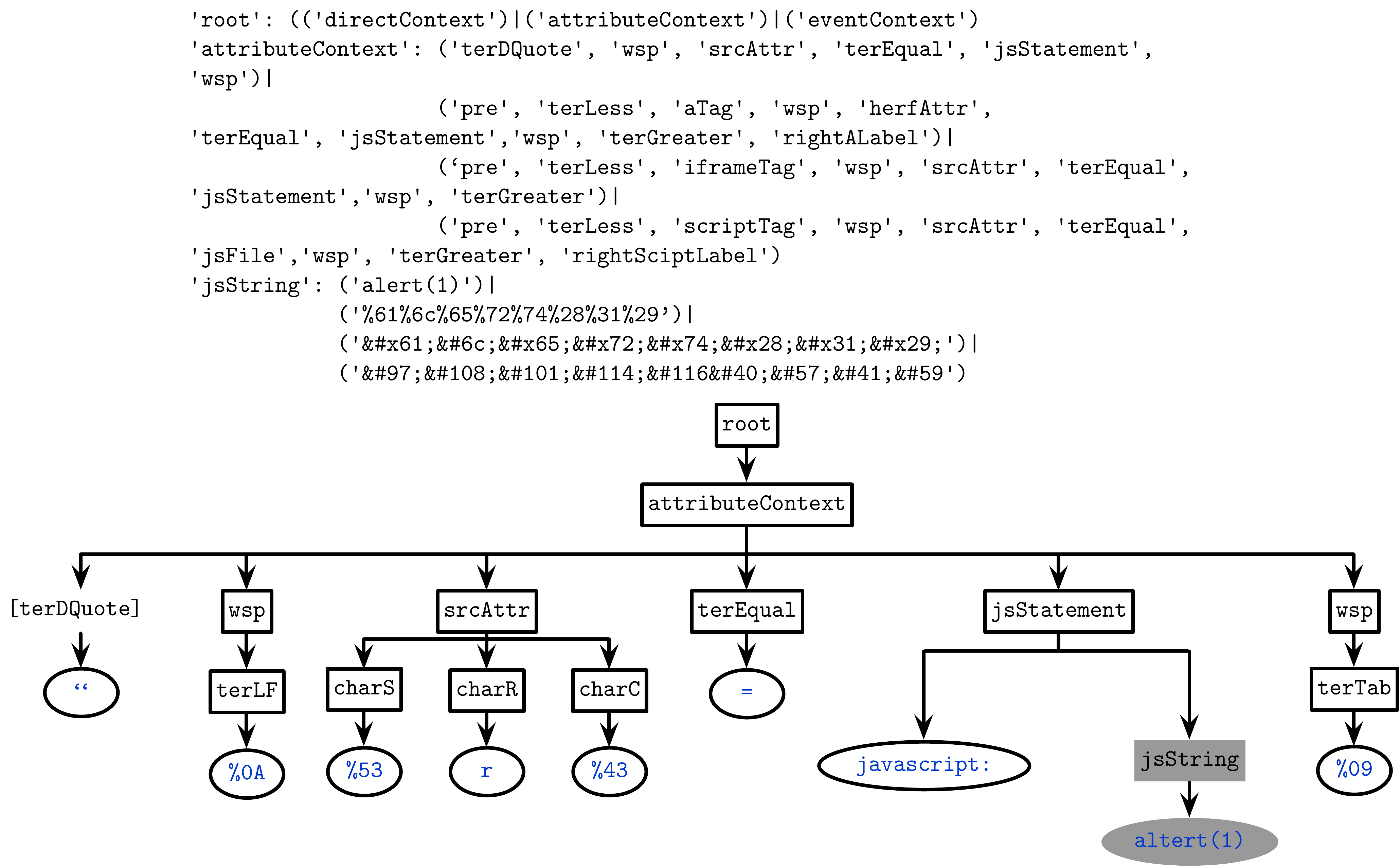}
	\caption{An illustrative example of the grammar tree for XSSi.}
	\label{fig:xss_grammar}
\end{figure*}

To have an intuitive illustration of the CFG and its corresponding test input generation mechanism, \pref{fig:xss_grammar} gives a sample grammar of XSSi while the complete grammar can be found from our supplementary document\footnote{~\href{https://tinyurl.com/3cu4ahrb}{https://tinyurl.com/3cu4ahrb}}. In \pref{fig:xss_grammar}, \textcolor{blue}{\texttt{\lq$\rightarrow$\rq}} represents production; \textcolor{blue}{\texttt{\lq,\rq}} represents connector; \textcolor{blue}{\texttt{\lq|\rq}} represents replacement sign. In order to generate a test input, we start from the root and the branches of the grammar tree that are generated according to a some predefined rules of the grammar in a random manner till the leaf node is reached. At the end, the combination of leaf nodes constitutes a test input. It is worth noting that the grammar is generic. Let us look at the test input example shown in~\pref{fig:xss_grammar} again. \textcolor{blue}{\texttt{\%0A\%53r\%43=javascript:alert(1)\%09}} gives a popup attack by using \textcolor{blue}{\texttt{alert(1)}}. In particular, \textcolor{blue}{\texttt{alert(1)}} is derived from a non-leaf node \texttt{jsString}. Therefore, the other variants of \textcolor{blue}{\texttt{alert(1)}} can also be included in the generator of \texttt{jsString}. Note that a simple alternation of content like \textcolor{blue}{\texttt{alert(2)}} does not come up to a feasible variant while only some semantically similar variants (e.g., \textcolor{blue}{\texttt{\%61\%6c\%65\%72\%74\%28\%31\%29,\&\#x61}}) count.

\subsubsection{Datasets Construction}
\label{sec:dataset_construction}

Based on the test inputs generated from the CFG of all the types of injection attack considered in \our, we can build the dataset for training our surrogate classifiers and the multi-task translation model as well as the initial seeds for the test input generation in MTEA thereafter. Note that it is not uncommon that there is a readily available dataset for injection testing of a given type of injection attack from past releases\footnote{~\url{https://github.com/payloadbox}}, which can be of great help to enrich the dataset. To demonstrate a wide applicability of \our, this paper assumes that there is no readily available dataset. 

Note that although using the CFG alone can search for injection cases, its capability is rather limited. According to our preliminary experiments, the number of bypassing injection cases, as well as the diversity, generated by the CFG alone is not sufficient to train a capable surrogate classifier and neural translation models, even when being allocated with a sufficiently large amount of computational budget. In addition, it highly depends on the type of injection attack. In contrast, \our\ has shown an outstanding performance for injection case generation in terms of both successful rate and the diversity.

\subsection{Surrogate Classifiers Learning with Word Embedding}

For most existing WAFs, a test input can only result in a binary outcome, i.e., either pass or fail. This does not provide sufficient information to evaluate the effectiveness of a test input and can, even worse, mislead the search-based test input generation in \our. To address this problem, we train a surrogate classifier for each type of injection attack by using the data collected in~\pref{sec:grammar} as a priori. As reported in a recent study~\cite{LiuLC20}, semantic information of test inputs can significantly improve the injection testing. In this work, we treat the test inputs of a type of injection attack akin to sentences from a natural language. Then, we leverage techniques from the NLP domain to build the language model for each type of injection attack. In the following paragraphs, we first introduce the language model used in \our\ and then describe the mechanism of the surrogate classifier.

\subsubsection{Word Embedding} 
\label{sec:word_embedding}

As mentioned in Section~\ref{sec:bg-nlm}, to better handle the semantic knowledge in the test inputs, the first step is to convert a sequence of words belonged to a test input into a word vector and this is also critical for \our\ to understand the semantics embedded in test inputs. For example, the language model needs to be able to understand that \textcolor{blue}{\texttt{\lq+\rq}}, \textcolor{blue}{\texttt{\lq/**/\rq}}, \textcolor{blue}{\texttt{\lq\%20\rq}}, \textcolor{blue}{\texttt{\lq\%09\rq}} are synonyms for the blank character in a test input. They merely represent different forms that disguise the attacks.

More specifically, we apply the classic \texttt{Word2vec} in \our\ to train a NLM that converts a test input into the corresponding word vectors. In particular, we use the continuous bag of words (CBOW) model~\cite{MikolovSCCD13}, which is a context aware version of \texttt{Word2vec}, to identify similar words in test inputs based on the contextual information, as the meaning of each single word in a test input vary depending on the context. Suppose that $w_i$ is the $i$-th word in a test input, the CBOW model correlates a target word $w_i$ and its context words under a given window size. For example, if the window size is two, the context words are $(w_{i-2},w_{i-1},w_{i+1},w_{i+2})$. Note that the context information have shown to be promising on providing more accurate vector embedding of the test inputs~\cite{LiuLC20}.

\subsubsection{Surrogate Classifier}
\label{sec:classifier}

There are many classifiers~\cite{PinzonPHCBC13,MakiouBS14} available for predicting whether an injection test input can bypass the WAF or not. In this work, we choose three neural network models, i.e., long-short term memory (LSTM)~\cite{HochreiterS97} network, recurrent neural networks (RNN)~\cite{GreffSKSS17} and gated recurrent unit (GRU)~\cite{ChoMGBBSB14} to serve our purpose. There are four major reasons.

\begin{itemize}
    \item They have been reported to be highly effective in handling the semantics of languages, i.e., they work well with the vector embedding of words~\cite{CollobertW08,MikolovCCD13}.
    \item They have been widely used in prior works~\cite{Kim14,SocherLNM11,SocherPWCMNP13}.
    \item It is plausible to build a large set of data samples since the possible number of test inputs for injection testing is enormously high.
    \item They only need to be trained once at the beginning and thus the training overhead is acceptable.
\end{itemize}

In practice, the classifier\footnote{It is worth noting that the validation accuracy of surrogate classifier is approximately range from 90 to 99\% in our offline experiments. Based on context-free grammar developed in~\pref{sec:grammar}, our training dataset consists of diverse types of test inputs. These are adequate to support the effectiveness of surrogate classifier.} takes the vector embedding of a test input as input and predicts a probability that decides whether this test input can bypass the WAF or not. In \our, we leverage this output probability as the fitness function to guide the search-based test input generation.

\subsection{Multi-task Translation Framework}
\label{sec:translation_model}

In \our, the translation between the test inputs for different types of injection attack is conceptually similar to the cross-lingual translation in NLP. This is the foundation of knowledge transfer in \our. It aims to translate a high quality test input for one type of injection attack to a semantically related and meaningful test input for another type. Such translation enables \our\ to share and unify knowledge among different types of injection attack. More specifically, there are two main steps as follows.

\subsubsection{Data Preprocessing}
\label{sec:preprocessing}

Generally speaking, the input of our multi-task translation model is a test input for one type of injection attack, while the output is the \lq translated\rq\ test input for another type. In this case, a data instance is a pair of semantically related test inputs from two types of injection attack, and the translation is asymmetric. Henceforth, when there are, say, six different types of injection attack, we need to prepare $\binom{6}{2}=15$ pairs of datasets and models to cover all bidirectional translations.

Given a pair of datasets for two types of injection attack, it is challenging to evaluate the semantic similarity between two test inputs. In \our, we use the latent semantic indexing model (provided by Gensim\footnote{\url{https://radimrehurek.com/gensim}}) to measure the similarity of semantic correlation between test inputs from two different types of injection attack. By doing so, any test input for one type of injection attack is paired with a (most similar) test input for another type. Thereafter, the paired data is stored into a dataset.

\subsubsection{Multi-Task Injection Translation}
\label{sec:seq2seq}

\begin{figure}[t!]
	\centering
	\includegraphics[width=1\linewidth]{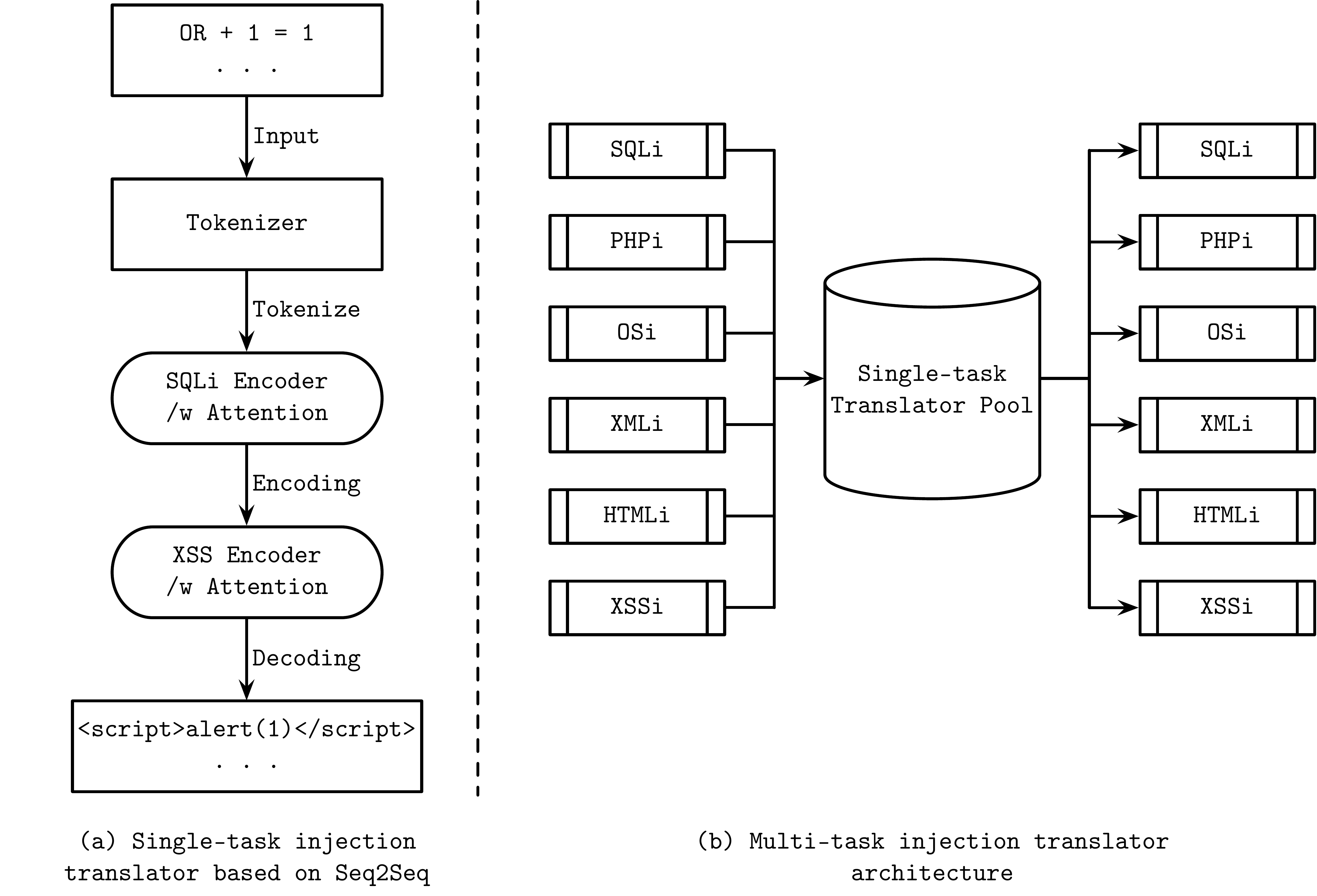}
	\caption{Illustrative example of the \texttt{Seq2Seq} framework for test input translations.}
	\label{fig:seq2seq}
\end{figure}

To capture the semantic information of a type of injection attack, \our\ chooses \texttt{Seq2Seq}~\cite{BahdanauCB14}, a LSTM-based \texttt{Seq2Seq} translation model\footnote{Our translation models are built upon \texttt{\href{https://opennmt.net}{OpenNMT-Py}}, a popular open source neural machine translation system.} to serve the purpose. Note that \texttt{Seq2Seq} has been widely used in many downstream NLP tasks including machine translation~\cite{BahdanauCB14}, text summarization~\cite{PaulusXS18}, and question answering~\cite{Al-RfouPSSSK16,ShangLL15}, to name a few. \texttt{Seq2Seq} is able to transform a sequence of words into another relevant sequence, thus enabling the translation of test inputs from one type of injection attack into another. Let us consider the example shown in~\pref{fig:seq2seq}(a), the incoming test input is \textcolor{blue}{\texttt{OR+1=1}} while its corresponding outcome could be \textcolor{blue}{\texttt{<scrpit>alert(1)</script>}}.

As shown in~\pref{fig:seq2seq}(a), \texttt{Seq2Seq} model consists of two key components, i.e., attention-based \texttt{encoder} and \texttt{decoder}, each of which can be regarded as an independent LSTM model~\cite{GreffSKSS17}. The inputs of the \texttt{encoder} are a sequence of word vectors $X=\{\mathbf{x}_1,\cdots,\mathbf{x}_s\}$ while the outputs of the \texttt{decoder} are another sequence of word vectors $Y=\{\mathbf{y}_1,\cdots,\mathbf{y}_p\}$. In particular, $s$ and $p$ respectively indicates the number of characters of the input and output sequences.

In a nutshell, \texttt{Seq2Seq} model aims to learn the following conditional distribution based on which the prediction of a sequence can be made: 
\begin{equation}
p\left(\mathbf{y_{1}},\cdots,\mathbf{y_{p}}|\mathbf{x_{1}},\cdots, \mathbf{x_{s}}\right).  
\end{equation}
Specifically, each word of the target sequence is conditioned on the following conditional probability:
\begin{equation}
p\left(\mathbf{y}_{t+1} | \mathbf{y}_1,\cdots, \mathbf{y}_t\right)=g\left(\mathbf{h}_{y_t},\mathbf{c}_{t+1}, \mathbf{y}_t\right).
\end{equation}

Given the input sequence $\{\mathbf{x_1},\cdots,\mathbf{x_s}\}$, we can calculate and update the hidden states of the \texttt{encoder} as $\{\mathbf{h_{x_1}},\cdots, \mathbf{h_{x_s}}\}$. Then, we use the attention mechanism~\cite{BahdanauCB14} to calculate the context vector $C=\{\mathbf{c}_1,\cdots, \mathbf{c}_p\}$, where $\mathbf{c}_i=\sum_{t=1}^{s} a_{i_t} \mathbf{h}_t$ and weight $a_{i_t}$ enables the \texttt{Seq2Seq} to focus on different parts of the input sequence when predicting the word vector $\mathbf{y}_i$. At the end, the \texttt{decoder} sequentially predicts the target word by using the hidden states and the context vector $C$ which also serves as the initial hidden state of the \texttt{decoder}. Note that the hidden states will be updated as words being generated. This process stops when the last word in the sequence is predicted.

In \our, a test input from one type of injection attack can in principle be translated into any other type. This cross-lingual translation is implemented by a multi-task learning paradigm, as shown in~\pref{fig:seq2seq}(b), where we train the pair-wise translators (i.e., the single-task translator based on \texttt{Seq2Seq} as shown in~\pref{fig:seq2seq}(a)) for any two of the injection types. Our preliminary experiments show that the single-task translator outperforms the parallel multi-task translation architecture. This structure enables the semantic information to be shared between the test input pairs for any two types of injection attack during the training process. Thus, it can produce more effective translation models.

\subsection{Multi-Task Evolutionary Test Input Generation}
\label{sec:mtea}

In \our, we develop a MTEA, extended from the classic single-task $(\mu+\lambda)$ evolutionary algorithm where $\mu=\lambda$, to evolve and automate the test input generation across multiple types of injection attack. There are four main reasons to use evolutionary algorithm to facilitate the test input generation~\cite{ChenLTL22}.

\begin{itemize}
    \item The search space of test inputs is rather complex and unknown a priori. This black-box nature makes evolutionary algorithm perfectly suitable.
    \item The possible number of viable test inputs for a type of injection attack is too large to enumerate~\cite{LiuLC20}.
    \item Due to the population-based nature, an evolutionary algorithm has a promising potential to maintain a promising diversity. Thus, it can be helpful for test input generation~\cite{AppeltNPB18,JanPAB19,LiuLC20}.
    \item Due to the iterative nature of evolutionary algorithm, it can be relatively slow to obtain a well converged solution. This is nonetheless less important than the diverse test inputs that can be produced by the evolutionary algorithm.
\end{itemize}

As the pseudo-code shown in \pref{alg:mtea}, each type of injection attack has its own population of test inputs, which can be shared with each others during the search. To comply with the syntactical correctness, the initial population of each injection type is fully seeded by the test inputs generated according to the CFGs introduced in~\pref{sec:grammar} (lines $1$ to $4$). To improve the evolution efficiency, the bypassing cases are removed before evaluation in every generation.

Note that at each generation, we run the test inputs in all the populations against the WAF, after which those can successfully bypass are removed from its population but stored in the corresponding archive for later evaluation (lines $19$ to $23$). Moreover, the individual injection case whose fitness is below the average fitness of the population is replaced by randomly generated injections in order to maintain a promising balance between convergence and diversity at the early stage of evolution. This can help avoid the search being trapped at some patterns of injection attacks that have already discovered the relevant vulnerabilities. This is important for injection testing~\cite{AppeltNPB18}.

\begin{algorithm}[t!]
    \DontPrintSemicolon
    \footnotesize
    \caption{Multi-Task Evolutionary Algorithm.}
    \label{alg:mtea}
	\KwIn{
	\\
	~~~~$\mathcal{G}$: context-free grammars for all injection tasks\\
	~~~~$\mathcal{C}$: surrogate classifiers for fitness assignment\\
	~~~~$\mathcal{T}$: multi-task translation models}
    \KwOut{
    \\
    ~~~~$\mathcal{A}$: bypassing cases for all injection tasks}
    
    

    \ForEach{$\mathcal{P}_i\in\mathcal{P}$}  
    {
       $\mathcal{P}_i\leftarrow$\textsc{genearteTestInputs($\mathcal{G}_i \in \mathcal{G}$)} \footnotemark{10}
       \\
       \tcc*[l]{Sharing mating pool composing of all populations}
       $\mathcal{M}_i\leftarrow\mathcal{P}_i$, $\mathcal{M}_i \in \mathcal{M}$\\
    }
    \While{The computational budget is not exhausted}
    {
      \ForEach{$\mathcal{P}_i\in\mathcal{P}$}  
      {  
		$\mathcal{P'} \leftarrow \emptyset$ \tcc*[r]{Offspring injections in $\mathcal{P}_i$}
        $\mathcal{B} \leftarrow \emptyset$ \tcc*[r]{Bypassed injections in $\mathcal{P}_i$}
        \ForEach{$\mathcal{S}\in\mathcal{P}_i$}
        { 
        	${S}_x \leftarrow \textsc{getFromMatingPool($\mathcal{M}_x \in \mathcal{M}$)}$\footnotemark{}\\
			$\mathcal{S}_t\leftarrow$\textsc{translate($\mathcal{S}_x, \mathcal{T}_k \in \mathcal{T}$)}\\
			\uIf{\textsc{bypassingWAF}($\mathcal{S}_t$)\footnotemark{} == True}
			{
				$\mathcal{B}\leftarrow\mathcal{S}_x$
			}
	        \Else{
    		    $\mathcal{{S}}_m\leftarrow$\textsc{mutate($\mathcal{S}$)}\\
		    	\If{\textsc{bypassingWAF}($\mathcal{S}_m$)==True}
				{
					$\mathcal{B}\leftarrow\mathcal{S}_t$
				}
	        }
	        $\mathcal{P'}\leftarrow\mathcal{P'}\bigcup(\mathcal{S}_x|\mathcal{S}_m$)\\
        }
        
      	$\mathcal{A}_i\leftarrow \mathcal{A}_i \bigcup \mathcal{B}$\\ 
		$\mathcal{P'}\leftarrow \mathcal{P'} - \mathcal{B}$\\  
        \textsc{fitnessEvaluate($\mathcal{P'}, \mathcal{C}_i \in \mathcal{C}$)}\footnotemark{}\\
        $\mathcal{U}\leftarrow$\textsc{sortByFitness($\mathcal{P}_i\bigcup\mathcal{P'}$)}\footnotemark{}\\
		$\mathcal{P}_i\leftarrow$top $m$ test inputs from $\mathcal{U}$\\  
      }
      \ForEach{$\mathcal{P}_i\in\mathcal{P}$}  
      {  
        \tcc*[l]{Update the sharing mating pool}
        $\mathcal{M}_i\leftarrow \mathcal{P}_i$, $\mathcal{M}_i \in \mathcal{M}$\\ 
      }
    }
    \Return $\mathcal{A}$
\end{algorithm}

\footnotetext[10]{Return a population generated using the CFG} 
\footnotetext[11]{Return a randomly selected injection to be a translated} 
\footnotetext[12]{Return the result of bypassing check} 
\footnotetext[13]{Assign fitness using corresponding surrogate classifiers} 
\footnotetext[14]{Return the sorted population according to individual's fitness} 

\subsubsection{Representation}
\label{sec:representation}

Given the support of the word embedding in the surrogate classifiers and translation models for each type of injection attack, our encoding for the MTEA in \texttt{\textsc{DaNuoYi}} is a string of words, representing a particular test input that can be of a different length. As such, the evolved test inputs can be directly attached to the HTTP request sent to the WAF for performance evaluation.

\subsubsection{Objective Function}
\label{sec:objective_function}

As discussed in~\pref{sec:classifier}, the trained surrogate classifiers serve as the fitness function that evaluates the likelihood of a test input bypassing the WAF (line $21$). Since such a classifier is empowered by the word embedding, no additional processes are required for evaluating a test input.

\subsubsection{Mutation Operators}
\label{sec:mutation}

To better maintain syntactically correct test inputs, \texttt{\textsc{DaNuoYi}} only uses mutation operators for offspring reproduction (line $15$). In particular, we develop the following six word-level mutation operators to mutate a test input into another different yet semantically related test input. In practice, at least one mutation operator will be randomly selected with the same probability for offspring reproduction:
\begin{enumerate}
	\item \textit{\underline{Grammar tree transformation:}} Based on the CFGs defined in \pref{sec:grammar}, it transforms a subtree of the grammar tree into another format. For example, \textcolor{blue}{\texttt{jsString->} \texttt{alert(1)}} to \textcolor{blue}{\texttt{\%61\%6c\%65\%72\%74\%28\%31\%29}}
	
	\item \textit{\underline{Text transformation:}} It confuses the capital and small letters in a test input. For example, \textcolor{blue}{\texttt{SCRIPT}} to \textcolor{blue}{\texttt{sCrIPT}}.
	
	\item \textit{\underline{Blank replacement:}} It replaces the blank character in a test input with an equivalent symbol. For example, \textcolor{blue}{\texttt{"+alert('XSSi')+"}} to \textcolor{blue}{\texttt{"\%20alert('XSSi')\%20"}}.
	
	\item \textit{\underline{Comment concatenation:}} It randomly adds comments between two words. For example, \textcolor{blue}{\texttt{<table background=\lq \rq></table>}} to \textcolor{blue}{\texttt{<table/*inj-}} \textcolor{blue}{\texttt{ection*/background=\lq \rq></table>}}.
	
	\item \textit{\underline{ASCII mutation:}} It mutates a word to its equivalent ASCII encoding format.
	
	\item \textit{\underline{Unicode mutation:}} It mutates a word to its equivalent Unicode format.
\end{enumerate}
The first mutation operator, i.e., \textit{Grammar tree transformation}, depends on the underlying injection type, as it is based on the corresponding CFG which is language specific; while the other five mutation operators are generic. In particular, all mutation operators are syntax-compliant since they either make change in a way that complies with the corresponding grammar as the first operator or perform straightforward encoding amendment and comment insertion, etc, as the other five operators.

\subsubsection{Knowledge Sharing and Multi-Task Evolution}

As shown in Algorithm~\ref{alg:mtea}, to facilitate the evolution of test input generation across multiple types of injection attack from different populations, the MTEA in \texttt{\textsc{DaNuoYi}} carries out a multi-task evolution in which each task is a standard evolutionary algorithm that evolves the test inputs with respect to a given type of injection attack (lines $5$ to $26$). Different from the traditional single-task evolutionary algorithm, in which the mating parents are merely from the same parent population, the offspring generation in MTEA aims to take advantages of elite information from all tasks. In particular, MTEA maintains a sharing mating pool, in which the mating parents are collected from the populations of any randomly chosen tasks by using a
$top$-$k$ ($k$ equals the population size) fitness ranking selection mechanism (line $22$). Such a mating pool is updated at the end of every generation (lines $24$ to $26$). Since the test input generation for one type of injection attack can exploit the semantic knowledge of the promising test inputs from other relevant types, we can expect to have a better chance to find more sophisticated test inputs for injection. In particular, the translation of a test input from one type of injection attack to another is realized by a
corresponding translation model as introduced in~\pref{sec:translation_model}. Note that if the translation fails, a mutation operation will be conducted to amend this translated test input. Note that since the initial population is seeded by test inputs generated according to different CFGs, which are malicious by themselves, we can expect that the test inputs generated by MTEA are still malicious.


\section{Experiments and Evaluation}
\label{sec:evaluation}

In this section, we evaluate and analyze the effectiveness of \our\ through answering the following research questions (RQs) in the presence of multiple types of injection attack.
\begin{itemize}
	\item \textbf{\underline{RQ1}:} Does \our\ find more valid test inputs that bypass the WAF than its single-task counterparts respectively designed for each type of injection attack? Note that the corresponding single-task counterpart does not apply the translation between different types of injection attack.
	\item \textbf{\underline{RQ2}:} Does \our\ find more valid test inputs than a random search (based on grammar only) with translation but without the mutation operators? 
	\item \textbf{\underline{RQ3}:} Does \our\ find as many bypassed injection attacks as the variant without fitness ranking? Note that this variant applies the translation but it does not have an elitism.
	\item \textbf{\underline{RQ4}:} If positive results are witnessed in \textbf{RQ1} and \textbf{RQ2}, can we interpret why does \our\ work?
	\item \textbf{\underline{RQ5}:} What happens if we merely rely on the use of the dedicated CFG to generate test inputs?
	
\end{itemize}
All experiments were carried out on a server equipped with $64$-bit Cent OS7, which has two Intel Xeon Platinum 8160 CPU ($48$ Cores $2.10$GHz), $256$GB RAM and four RTX 2080 Ti GPUs. The target WAFs were deployed on a virtual cloud server with $64$-bit Ubuntu 18.04, which is based on the AMD EPYC 7K62 $48$-Core CPU and has $1$GB RAM.
\begin{remark}
    \textbf{RQ1} seeks to evaluate the effectiveness and the added value of our proposed multi-task test input generation paradigm.
\end{remark}
\begin{remark}
    \textbf{RQ2} aims to evaluate the effectiveness of mutation operators in the evolutionary algorithm.
\end{remark}
\begin{remark}
    \textbf{RQ3} aims to evaluate the importance of the selection pressure in evolutionary algorithm.
\end{remark}
\begin{remark}
    \textbf{RQ4} plans to analyze the working mechanism of \our\ by investigating the semantic knowledge learned by the language model.
\end{remark}
\begin{remark}
    \textbf{RQ5} aims to the investigate the potential drawbacks of merely using the CFG-based method for the test input generation. In addition, it also plans to understand the bottlenecks of \our.
\end{remark}

\subsection{Experiment Setup}
\label{sec:setup}

\subsubsection{Types of Injection Attack} In theory, \our\ can generate any type of injection attacks to test the vulnerabilities of the system under test as long as the computational budget permits. In this work, we examine six types of injection attack chosen according to their prevalence and severity\footnote{~\href{https://www.akamai.com/us/en/multimedia/documents/state-of-the-internet/soti-security-web-attacks-and-gaming-abuse-executive-summary-2019.pdf}{https://www.akamai.com/}}, as discussed below.

\begin{itemize}
    \item\underline{SQL injection (SQLi):} This is the most prevalent injection attack on relational database management systems. SQLi usually leads to severe data breaches.
    \item\underline{XSS injection (XSSi):} This injection is mainly related to Javascript, the vulnerabilities of which allow an attacker to inject malicious web links.
    \item\underline{XML (XPath) injection (XMLi):} This injection includes compromising the data, it encodes and causes its query language ---XPath--- to result in severe data breaches.
    \item\underline{HTML injection (HTMLi):} This injection can result in a modification of a web page rendered by the web application, which potentially affects all its visitors.
    \item\underline{OS command injection (OSi):} Shell script is often hard-coded in a web application. Relevant injection vulnerabilities can allow malicious system-level commands to be injected, such as listening on a port or a fork bomb.
    \item\underline{PHP injection (PHPi):} This may cause path traversal or denial-of-service attacks, depending on the context.
\end{itemize}

\subsubsection{Subject WAFs}
\label{subject_waf}

To evaluate the practicality and improve the external validity, we evaluate \our\ on the following three widely used real-world WAFs.

\begin{itemize}
    \item \texttt{ModSecurity}\footnote{~\url{https://www.modsecurity.org}}: This is a cross-platform WAF and it serves as the fundamental security component for \texttt{Apache HTTP Server}, \texttt{Microsoft IIS}, and \texttt{Nginx}, which underpin millions of the web applications worldwide. This WAF maintains a large amount of rule sets that provide defense mechanisms for various types of injection attack.
    \item \texttt{Ngx-lua-WAF}\footnote{~\url{https://github.com/loveshell/ngx\_lua\_waf}}: This is a scalable WAF designed for high-performance web applications. It also supports the defense on different types of injection attack with rule sets complement those covered by \texttt{ModSecurity}.
    \item \texttt{Lua-resty-WAF}\footnote{~\url{https://github.com/p0pr0ck5/lua-resty-waf}}: This is another popular and scalable open-source WAF based on \texttt{OpenResty}\footnote{\url{https://openresty.org}}. It supports the patterns extended from \texttt{ModSecurity}.
\end{itemize}

We deploy a dummy web application behind the WAFs. Note that in this work, the testing target is the WAF; the web application merely serves as the destination of the attacks.

\subsubsection{Dataset}
\label{sec:dataset}

As discussed in~\pref{sec:grammar}, we use the CFGs to generate the initial test inputs which constitute the datasets for training both the surrogate classifiers and the multitask translation framework of \our\ shown in~\pref{fig:seq2seq}. 

To train the classifier for each type of injection attack, we generate $20,000$ test inputs for the underlying WAF. They are labeled as either \textit{blocked} or \textit{bypassed} according to the results when feeding the test inputs into the WAF. To train the models for the bidirectional translation between any two types of injection attacks as introduced in~\pref{sec:preprocessing}, we need $\binom{6}{2}=15$ pairs of datasets given six types of injection attack. To that end, we generate $30,000$ pairs of translatable test inputs for each of these $15$ pairs of datasets.

\subsubsection{Settings}
\label{sec:parameter_settings}

\our\ synergies several key techniques of NLP and search-based software testing, each of which has some hyper-parameters. In our empirical study, the relevant hyper-parameters are set as according to the results of offline parameter tuning.
\begin{itemize}
	\item\underline{Word2Vec:} The dimension of the embedding vector of each token w.r.t. a test injection in \our\ as $128$.
	\item\underline{Classifier}: For each type of injection attack, we apply three different neural networks (i.e., RNN, LSTM and GRU) as the alternative classifiers. They are all set to have one hidden layer. All the classifiers share the same pretrained \texttt{Word2Vec} embedding.
	\item\underline{Translation Model}: The LSTM under the \texttt{Seq2Seq} framework is set with $128$ hidden units, which are consistent to the surrogate classifiers.
	\item\underline{MTEA}: Each task maintains a population of $100$ solutions and the number of generations is set as $50$. Note that we find that these settings can strike a well balance between performance and efficiency.
\end{itemize}

\subsubsection{Metric and Statistical Test}
\label{sec:metrics}

To evaluate the ability for generation injection instances, we use the number of different bypassing test inputs as the performance metric. If a distinct test input bypasses the WAF, it means the identification of a potentially new vulnerability~\cite{AppeltNPB18}. To mitigate potential bias, each experiment is repeated $21$ independent runs with different random seeds. To have a statistical interpretation of the significance of comparison results, we use the following three statistical tests in our empirical study.
\begin{itemize}
	\item\underline{Wilcoxon signed-rank test}~\cite{Haynes2013}: This is a non-parametric statistical test that makes no assumption about the underlying distribution of the data. In particular, the significance level is set to $p=0.05$ in our experiments.
    \item\underline{Scott-Knott test}~\cite{MittasA13}: Instead of merely comparing the raw metric values, we apply the Scott-Knott test to rank the performance of different peer techniques over $21$ runs on each test scenario. In a nutshell, the Scott-Knott test uses a statistical test and effect size to divide the performance of peer algorithms into several clusters. The performance of peer algorithms within the same cluster are statistically equivalent. Note that the clustering process terminates until no split can be made. Finally, each cluster can be assigned a rank according to the mean metric values achieved by the peer algorithms within the cluster. In particular, the larger the rank is, the better performance of the algorithm achieves.
    \item\underline{$A_{12}$ effect size}~\cite{VarghaD00}: To ensure the resulted differences are not generated from a trivial effect, we apply $A_{12}$ as the effect size measure to evaluate the probability that one algorithm is better than another. Specifically, given a pair of peer algorithms, $A_{12}=0.5$ means they are \textit{equivalent}. $A_{12}>0.5$ denotes that one is better for more than 50\% of the times. $0.56\leq A_{12}<0.64$ indicates a \textit{small} effect size while $0.64 \leq A_{12} < 0.71$ and $A_{12} \geq 0.71$ mean a \textit{medium} and a \textit{large} effect size, respectively. 
\end{itemize}
Note that both Wilcoxon signed-rank test and $A_{12}$ effect size are also used in the Scott-Knott test for generating clusters.

\subsection{Performance Comparison of \our\ with the Corresponding Single-Task Counterparts}
\label{sec:rq1}

\subsubsection{Methods}
\label{sec:methods_rq1}

To the best of our knowledge, \our\ is the first of its kind tool to generate test inputs for more than one type of injection attack simultaneously. To answer \textbf{RQ1}, we plan to compare our proposed \our\ with its single-task counterparts which can only generate test inputs for a given type of injection attack. There have been some tools to serve this purpose, such as \cite{AppeltNPB18} for SQLi, \cite{JanPAB19} for XMLi, and \cite{DucheneGRR12} for XSSi. Unfortunately, none of these tools are neither open source projects nor readily available. Therefore, we extract each task of the MTEA in \our\ as the peer method for automatic test input generation for a given type of injection attack (generally denoted as \texttt{STEA}). They serve as resemblances to the existing single-task tools given that they share similar learning and evolutionary search techniques. Since we use three different neural networks, i.e., RNN, LSTM, and GRU, as the surrogate classifiers, there are three different groups in the comparisons.

\begin{table*}[t!]
  \centering
  \caption{The total number of bypassed test inputs (over $21$ runs) obtained by CFG-based method and \texttt{STEA} versus \our\ with different surrogate classifiers under the same computational budget.}
  \footnotesize
    \centering
	\setlength{\tabcolsep}{1.0mm}{
    \begin{tabular}{c|c|c|c|c|c|c|c}
    \hline
    \multirow{1}[4]{*}{WAF} & \multirow{1}[4]{*}{Injection}  & \multicolumn{2}{c|}{RNN} & \multicolumn{2}{c|}{LSTM} & \multicolumn{2}{c}{GRU}\\
    \cline{3-8}      &       & \texttt{STEA}  & \our & \texttt{STEA}  & \our & \texttt{STEA}  & \our \\
    \hline
    \multirow{6}[2]{*}{\texttt{ModSecurity}} 
    & SQLi & 1798 (171.0)$^{\dag}$ & \bb{2232 (118.0)} & 1922 (93.0)$^{\dag}$ & \bb{2309 (102.0)} & 1844 (86.0)$^{\dag}$ & \bb{2127 (54.0)}\\
    & OSi  & 2622 (36.0)$^{\dag}$& \bb{3424 (18.0)} & 2555 (35.0)$^{\dag}$ & \bb{3375 (20.0)} & 2578 (25.0)$^{\dag}$ & \bb{3375 (30.0)} \\
    & PHPi & 4015 (105.0)$^{\dag}$ & \bb{4790 (13.0)} & 4111 (53.0)$^{\dag}$ & \bb{4815 (14.0)} &2578 (25.0)$^{\dag}$ & \bb{4800 (16.0)}  \\
    & XMLi  & 2010 (150.0)$^{\dag}$ & \bb{2337 (110.0)} & 1959 (155.0)$^{\dag}$ & \bb{2340 (103.0)} & 1876 (78.0)$^{\dag}$ & \bb{2188 (45.0)} \\
    & XSSi  & 1836 (99.0)$^{\dag}$ & \bb{2830 (125.0)} & 2435 (201.0)$^{\dag}$ & \bb{3100 (284.0)} & 1786 (169.0)$^{\dag}$ & \bb{2546 (95.0)} \\
    & HTMLi & 997 (693.0)$^{\dag}$ & \bb{2937 (102.0)} & 742 (234.0)$^{\dag}$ & \bb{2852 (306.0)} & 1122 (561.0)$^{\dag}$ & \bb{2773 (232.0)}  \\
    \hline\hline
    \multirow{6}[1]{*}{\texttt{Ngx-Lua-WAF}}
    & SQLi & 4610 (18.0)$^{\dag}$ & \bb{4984 (5.0)} & 4620 (16.0)$^{\dag}$ & \bb{4985 (6.0)} & 4602 (22.0)$^{\dag}$ & \bb{4984 (5.0)} \\
    & OSi   & 5000 (0.0) & 5000 (0.0) & 5000 (0.0) & 5000 (0.0) & 5000 (0.0) & 5000 (0.0) \\
    & PHPi  & 2434 (969.0)$^{\dag}$ & \bb{3471 (468.0)} & 2170 (716.0)$^{\dag}$ & \bb{3433 (114.0)} & 2156 (318.0)$^{\dag}$ & \bb{3617 (311.0)}\\
    & XMLi  & 4630 (26.0)$^{\dag}$ & \bb{4982 (6.0)} & 4625 (25.0)$^{\dag}$ & \bb{4984 (2.0)} & 4610 (12.0)$^{\dag}$ & \bb{4983 (4.0)} \\
    & XSSi  & 947 (37.0)$^{\dag}$ & \bb{1762 (46.0)} & 715 (63.0)$^{\dag}$ & \bb{1631 (50.0)} & 952 (72.0)$^{\dag}$ & \bb{1808 (60.0)}  \\
    & HTMLi & 1614 (180.0)$^{\dag}$ & \bb{3523 (56.0)} & 1744 (997.0)$^{\dag}$ & \bb{3599 (85.0)} & 2241 (166.0)$^{\dag}$ & \bb{3907 (105.0)}  \\
    \hline\hline
    \multirow{6}[1]{*}{\texttt{Lua-Resty-WAF}} 
    & SQLi & 2231 (124.0)$^{\dag}$ & \bb{2901 (118.0)} & 2338 (49.0)$^{\dag}$ & \bb{2965 (38.0)} & 2286 (94.0)$^{\dag}$ & \bb{2880 (50.0)} \\
    & OSi  & 4514 (17.0)$^{\dag}$ & \bb{4942 (9.0)} & 4469 (66.0)$^{\dag}$ & \bb{4936 (14.0)} & 4322 (128.0)$^{\dag}$ & \bb{4907 (9.0)}  \\
    & PHPi & 2789 (706.0)$^{\dag}$ & \bb{3967 (116.0)} & 2651 (736.0)$^{\dag}$ & \bb{4005 (53.0)} & 2372 (537.0)$^{\dag}$ & \bb{3918 (161.0)} \\
    & XMLi & 2373 (81.0)$^{\dag}$ & \bb{3034 (110.0)} & 2380 (138.0)$^{\dag}$ & \bb{2996 (78.0)} & 2288 (67.0))$^{\dag}$ & \bb{2935 (43.0)}  \\
    & XSSi & 1542 (85.0)$^{\dag}$ & \bb{3048 (219.0)} & 2560 (529.0)$^{\dag}$ & \bb{3717 (115.0)} & 1893 (233.0)$^{\dag}$ & \bb{3296 (111.0)}  \\
    & HTMLi & 1072 (843.0)$^{\dag}$ & \bb{3011 (136.0)} & 779 (256.0)$^{\dag}$ & \bb{2952 (221.0)} & 1158 (539.0)$^{\dag}$ & \bb{2771 (320.0)}  \\
    \hline
    \end{tabular}%
    }
    \begin{tablenotes}
    \item[1] Each cell shows the median value of the number of bypassing test inputs with the IQR value in the parentheses.
    \item[2] $^{\dag}$ denotes that \texttt{\textsc{DaNuoYi}} is significantly better than the peer algorithm according to the Wilcoxon rank-sum test at a $0.05$ significance level.
    \end{tablenotes} 
   
    \label{tab:rq1}
\end{table*}

\begin{figure*}[t!]
	\centering
	\includegraphics[width=\linewidth]{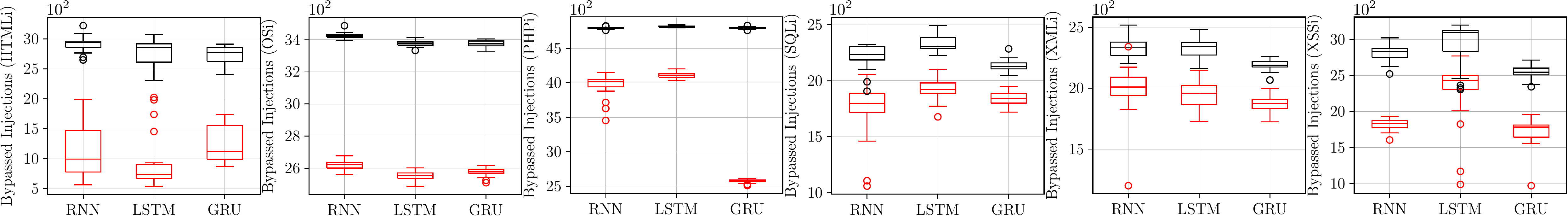}
	\caption{Box plots of the distributions of the bypassed test inputs generated by \our\ (denoted as the black boxes) compared against the corresponding single-task counterparts (denoted as the \textcolor{red}{red} boxes) with RNN, LSTM, GRU as the surrogate classifier, respectively, on \texttt{ModSecurity} over $21$ runs under the same computational budget. The numbers of bypassed injections are scaled by $10^2$.}
	\label{fig:box_mod}
\end{figure*}

\begin{figure*}[t!]
	\centering
	\includegraphics[width=\linewidth]{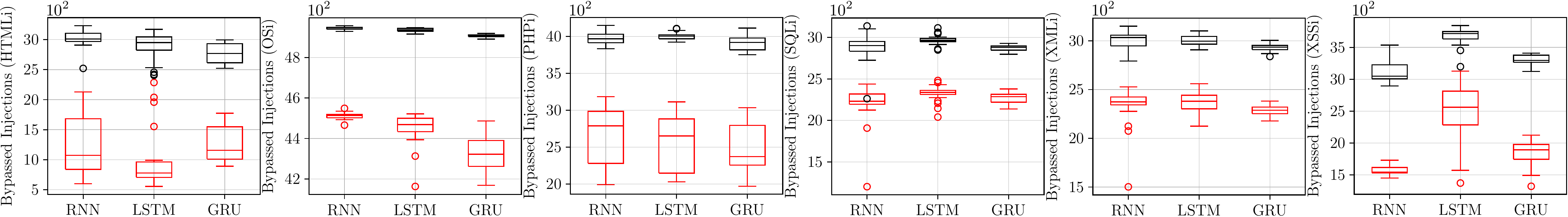}
	\caption{Box plots of the distributions of the bypassed test inputs generated by \our\ (denoted as the black boxes) compared against the corresponding single-task counterparts (denoted as the \textcolor{red}{red} boxes) with RNN, LSTM, GRU as the surrogate classifier, respectively, on \texttt{Lua-Resty-WAF} over $21$ runs under the same computational budget. The numbers of bypassed injections are scaled by $10^2$.}
	\label{fig:box_lua}
\end{figure*}

\begin{figure*}[t!]
	\centering
	\includegraphics[width=\linewidth]{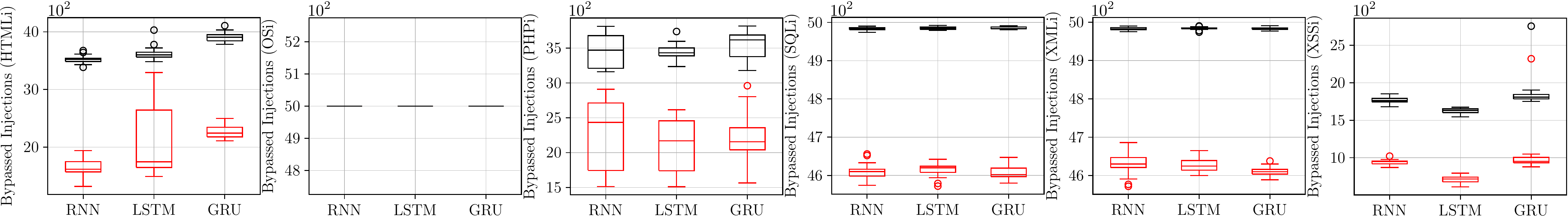}
	\caption{Box plots of the distributions of the bypassed test inputs generated by \our\ (denoted as the black boxes) compared against the corresponding single-task counterparts (denoted as the \textcolor{red}{red} boxes) with RNN, LSTM, GRU as the surrogate classifier, respectively, on \texttt{Ngx-Lua-WAF} over $21$ runs under the same computational budget. The numbers of bypassed injections are scaled by $10^2$.}
	\label{fig:box_ngx}
\end{figure*}

\begin{figure*}[t!]
	\centering
	\includegraphics[width=1\linewidth]{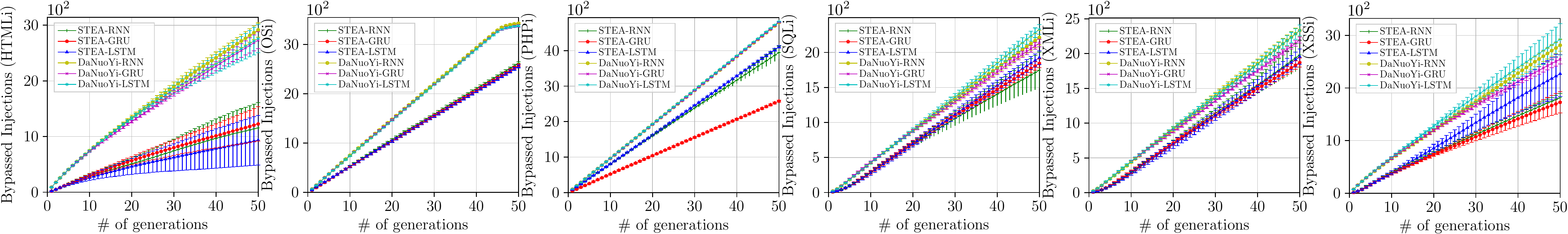}
		\caption{The number of valid test inputs generated by \our\ compared against the corresponding single-task counterparts with RNN, LSTM, GRU as the surrogate classifier, respectively, during the evolutionary process on \texttt{ModSecurity} over $21$ runs under the same computational budget (shown as error bars). The numbers of bypassed injections are scaled by $10^2$.}
	\label{fig:traj_mod}
\end{figure*}

\begin{figure*}[t!]
	\centering
	\includegraphics[width=1\linewidth]{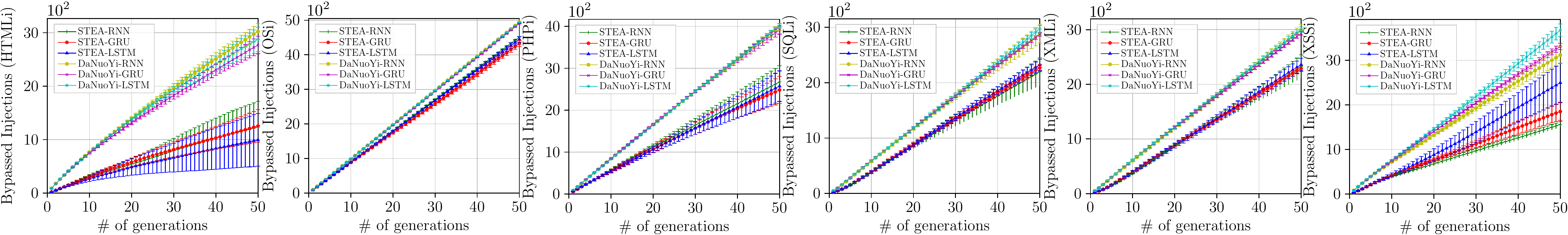}
	\caption{The number of valid test inputs generated by \our\ compared against the corresponding single-task counterparts with RNN, LSTM, GRU as the surrogate classifier, respectively, during the evolutionary process on \texttt{Lua-Resty-WAF} over $21$ runs under the same computational budget (shown as error bars). The numbers of bypassed injections are scaled by $10^2$.}
	\label{fig:traj_lua}
\end{figure*}

\begin{figure*}[t!]
	\centering
	\includegraphics[width=1\linewidth]{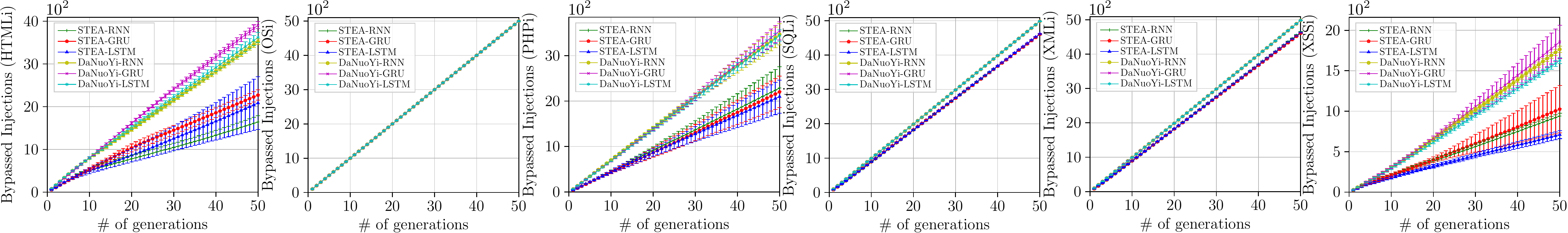}
	\caption{The number of valid test inputs generated by \our\ compared against the corresponding single-task counterparts with RNN, LSTM, GRU as the surrogate classifier, respectively, during the evolutionary process on \texttt{Ngx-Lua-WAF} over $21$ runs under the same computational budget (shown as error bars). The numbers of bypassed injections are scaled by $10^2$.}
	\label{fig:traj_ngx}
\end{figure*}

\begin{figure}[t!]
	\centering
	\includegraphics[width=.4\linewidth]{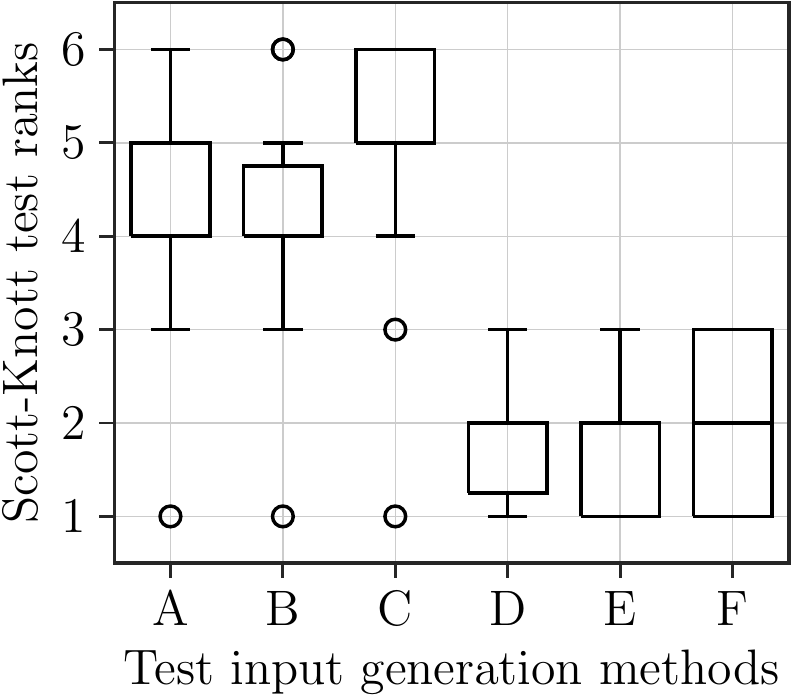}
	\caption{Box plots of Scott-Knott test ranks achieved by \our\ with RNN, LSTM, and GRU as the surrogate classifier, respectively, on all WAFs and Injection tasks compared against the corresponding STEAs. In particular, the x label from A to F represents \texttt{DaNuoYi-RNN}, \texttt{DaNuoYi-GRU}, \texttt{DaNuoYi-LSTM}, \texttt{STEA-RNN}, \texttt{STEA-GRU} and \texttt{STEA-LSTM}, respectively.}
	\label{fig:sk-rank}
\end{figure}

\begin{figure}[t!]
	\centering
	\includegraphics[width=.6\linewidth]{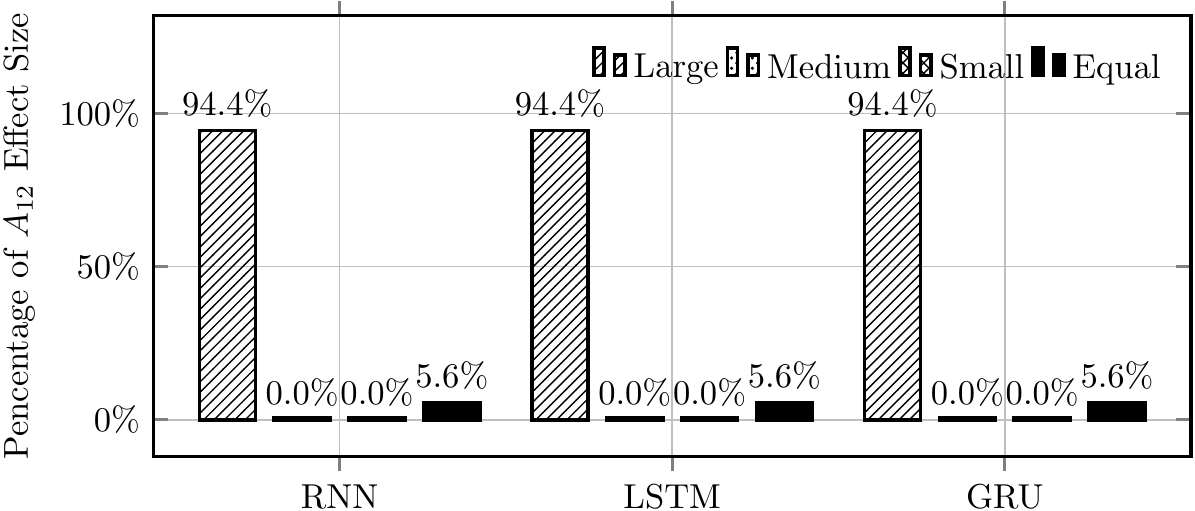}
	\caption{Percentage of the large, medium, small, and equal $A_{12}$ effect size, respectively, when comparing \our\ with the corresponding STEA that uses same surrogate classifier.}
	\label{fig:a12}
\end{figure}

\subsubsection{Results}
\label{sec:results_rq1}

\pref{tab:rq1} gives the comparison results of the total number of distinct test inputs, generated by different test input generation methods, bypass a given WAF under the same computational budget. According to the comparison results between \our\ and its corresponding single-task counterparts, it is clear to see that \our\ is able to find more bypassed test inputs (nearly up to $3.8\times$) for each type of injection attack on all three WAFs. The only exception is the generation of test inputs for the OSi on \texttt{Ngx-Lua-WAF} where all methods have shown constantly the same performance. This can be explained as the small search space in this scenario, which does not pose any challenge to the search of valid test inputs. As shown in~\pref{fig:box_mod} to~\pref{fig:box_ngx}, we apply the box plots to give a better statistical view upon the comparison results. From these figures, we can clearly see that the number of valid test inputs identified by \our\ is consistently larger than its single-task counterparts. 

In addition, to investigate the performance of \our\ against its corresponding single-task counterparts, we keep a record of the number of valid test inputs generated during the evolutionary process. As the trajectories shown in~\pref{fig:traj_mod} to~\pref{fig:traj_ngx}, it is clear to see that \our\ is able to generate more valid test inputs all the time. It is also interesting to note that these single-task test input generation methods can easily get stuck at the early stage of evolution; whereas the number of test inputs generated by \our\ steadily increases with the evolutionary process.

To have an overall comparison of \our\ against the corresponding single-task counterparts for all different types of injection attack w.r.t. all WAFs, we apply the Scott-Knott test upon the collected comparison results. From the box plots shown in~\pref{fig:sk-rank}, we find that \our\ have shown overwhelmingly better performance compared to the corresponding single-task counterparts.

\vspace{1em}
\noindent
\framebox{\parbox{\dimexpr\linewidth-2\fboxsep-2\fboxrule}{
    \textbf{\underline{Response to RQ1:}} \textit{From the empirical results discussed in this subsection, we confirm the effectiveness of \our. Specifically, by leveraging the similarity of semantic knowledge across different injection attacks, \our\ can generate more valid test inputs comparing against the corresponding single-task counterparts, which easily get stuck at the early stage of evolution. In addition, \our\ serves the purpose of generating multiple types of injection attacks in a multi-task manner.}}}

\subsection{Investigation of the Impacts of the Mutation Operators in \our}
\label{sec:rq2}

\begin{table*}[t!]
  \centering
  \caption{The total number of bypassed test inputs (over $21$ runs) obtained by \texttt{CFG-}\our, random search (\texttt{RAN}) and rule-based injection search (\texttt{RIS}) versus \our\ with different surrogate classifiers under the same computational budget. }
    \centering
    \resizebox{\linewidth}{!}{
	\setlength{\tabcolsep}{1.0mm}{
    \begin{tabular}{c|c|c|c|c|c|c|c||c|c}
    \hline
    \multirow{1}[4]{*}{WAF} & \multirow{1}[4]{*}{Injection}  & \multicolumn{2}{c|}{RNN} & \multicolumn{2}{c|}{LSTM} & \multicolumn{2}{c||}{GRU} & \multicolumn{1}{c|}{\multirow{1}[4]{*}{\texttt{RIS}}} & \multicolumn{1}{c}{\multirow{1}[4]{*}{\texttt{RAN}}}\\
    \cline{3-8}      &       & \texttt{CFG-}\our  & \our &\texttt{CFG-}\our  & \our & \texttt{CFG-}\our  & \our & &\\
    \hline
    \multirow{6}[2]{*}{\texttt{ModSecurity}} 
    & SQLi & 492 (24.0)$^{\dag}$ & \bb{2232 (118.0)} & 510 (17.0)$^{\dag}$ & \bb{2309 (102.0)} & 501 (16.0)$^{\dag}$ & \bb{2127 (54.0)} & 405 (23.0)$^{\dag}$ & 808 (65.0)$^{\dag}$\\
    & OSi  & 2167 (20)$^{\dag}$ & \bb{3424 (18.0)} & 2154 (33.0)$^{\dag}$ & \bb{3375 (20.0)} & 2168 (14.0)$^{\dag}$ & \bb{3375 (30.0)} & 2187 (10.0)$^{\dag}$ & 3182 (23.0)$^{\dag}$\\
    & PHPi & 3820 (30.0)$^{\dag}$ & \bb{4790 (13.0)} & 3815 (24.0)$^{\dag}$ & \bb{4815 (14.0)} & 3801 (13.0)$^{\dag}$ & \bb{4800 (16.0)}  & 3774 (23.0) $^{\dag}$ & 4662 (29.0)$^{\dag}$\\
    & XMLi  & 519 (20.0)$^{\dag}$ & \bb{2337 (110.0)} & 524 (13.0)$^{\dag}$ & \bb{2340 (103.0)} & 528 (12.0)$^{\dag}$ & \bb{2188 (45.0)} & 405 (23.0) $^{\dag}$ & 781 (75.0)$^{\dag}$\\
    & XSSi  & 1160 (35.0)$^{\dag}$ & \bb{2830 (125.0)} & 1202 (24.0)$^{\dag}$ & \bb{3100 (284.0)} & 1254 (36.0)$^{\dag}$ & \bb{2546 (95.0)} &849 (38.0)$^{\dag}$ & 1710 (88.0)$^{\dag}$\\
    & HTMLi & 1501 (37.0)$^{\dag}$ & \bb{2937 (102.0)} & 1517 (19.0)$^{\dag}$ & \bb{2852 (306.0)} & 1445 (31.0)$^{\dag}$ & \bb{2773 (232.0)}  & 967 (9.0)$^{\dag}$ & 1926 (67.0)$^{\dag}$\\
    \hline\hline
    \multirow{6}[1]{*}{\texttt{Ngx-Lua-WAF}}
    & SQLi & 4628 (11.0)$^{\dag}$ & \bb{4984 (5.0)} & 4618 (12.0)$^{\dag}$ & \bb{4985 (6.0)} & 4634 (29.0)$^{\dag}$ & \bb{4984 (5.0)}  & 4456 (25.0)$^{\dag}$ & 4960 (11.0)$^{\dag}$\\
    & OSi   & 5000 (0.0) & 5000 (0.0) & 5000 (0.0) & 5000 (0.0) & 5000 (0.0) & 5000 (0.0) & 5000 (0.0) & 5000 (0.0)\\
    & PHPi  & 2056 (19.0)$^{\dag}$ & \bb{3471 (468.0)} & 2066 (15.0)$^{\dag}$ & \bb{3433 (114.0)} & 2061 (29.0)$^{\dag}$ & \bb{3617 (311.0)} & 2074 (22.0)$^{\dag}$ & 3297 (47.0)$^{\dag}$\\
    & XMLi  & 4628 (6.0)$^{\dag}$ & \bb{4982 (6.0)} & 4621 (15.0)$^{\dag}$ & \bb{4984 (2.0)} & 4619 (25.0)$^{\dag}$ & \bb{4983 (4.0)} & 4456 (25.0)$^{\dag}$ & 4953 (13.0)$^{\dag}$\\
    & XSSi  & 1012 (26.0)$^{\dag}$ & \bb{1762 (46.0)} & 1010 (25.0)$^{\dag}$ & \bb{1631 (50.0)} & 1017 (26.0)$^{\dag}$ & \bb{1808 (60.0)} & 1175 (18.0)$^{\dag}$ & 2120 (86.0)\\
    & HTMLi & 2694 (36.0)$^{\dag}$ & \bb{3523 (56.0)} & 2667 (31.0)$^{\dag}$ & \bb{3599 (85.0)} & 2680 (15.0)$^{\dag}$ & \bb{3907 (105.0)}  &2746 (50.0)$^{\dag}$ & 4010 (67.0)\\
    \hline\hline
    \multirow{6}[1]{*}{\texttt{Lua-Resty-WAF}} 
    & SQLi & 1164 (25.0)$^{\dag}$  & \bb{2901 (118.0)} & 1131 (39.0)$^{\dag}$ & \bb{2965 (38.0)} & 1132 (48.0)$^{\dag}$ & \bb{2880 (50.0)}  & 1013 (19.0)$^{\dag}$ & 1864 (50.0)$^{\dag}$\\
    & OSi  & 4422 (7.0)$^{\dag}$  & \bb{4942 (9.0)} & 4423 (11.0)$^{\dag}$ & \bb{4936 (14.0)} & 4423 (16.0)$^{\dag}$ & \bb{4907 (9.0)}  & 4434 (8.0)$^{\dag}$ & 4916 (12.0)$^{\dag}$\\
    & PHPi & 2766 (22.0)$^{\dag}$  & \bb{3967 (116.0)} & 2773 (33.0)$^{\dag}$ & \bb{4005 (53.0)} & 2759 (40.0)$^{\dag}$ & \bb{3918 (161.0)} & 2767 (38.0)$^{\dag}$ & 3911 (54.0)$^{\dag}$\\
    & XMLi & 1136 (25.0)$^{\dag}$ & \bb{3034 (110.0)} & 1157 (49.0)$^{\dag}$ & \bb{2996 (78.0)} & 1135 (39.0)$^{\dag}$ & \bb{2935 (43.0)}  & 1013 (19.0)$^{\dag}$ & 1854 (60.0)$^{\dag}$\\
    & XSSi & 2067 (75.0)$^{\dag}$ & \bb{3048 (219.0)} & 2027 (46.0)$^{\dag}$ & \bb{3717 (115.0)} & 2063 (33.0)$^{\dag}$ & \bb{3296 (111.0)}  & 1817 (56.0)$^{\dag}$ & 3072 (49.0)$^{\dag}$\\
    & HTMLi & 1530 (25.0)$^{\dag}$ & \bb{3011 (136.0)} & 1529 (35.5)$^{\dag}$ & \bb{2952 (221.0)} & 1509 (46.0)$^{\dag}$ & \bb{2771 (320.0)}  & 1022 (17.0)$^{\dag}$ & 1994 (76.0)$^{\dag}$\\
    \hline
    \end{tabular}%
    }
    }
    \begin{tablenotes}
    \item[1] Each cell shows the median value of the number of bypassing test inputs with the IQR value in the parentheses.
    \item[2] $^{\dag}$ denotes that \texttt{\textsc{DaNuoYi}} is significantly better than the peer algorithm according to the Wilcoxon rank-sum test at a $0.05$ significance level.
    \end{tablenotes} 
   
    \label{tab:rq3}
\end{table*}

\subsubsection{Methods}
\label{sec:methods_rq2}

According to the experiments in~\pref{sec:results_rq1}, we have witnessed the effectiveness of our proposed \our\ for generating test inputs for different types of injection attacks. We argue that the six mutation operators developed in~\pref{sec:mutation} are the driving force to create valid test inputs for the underlying WAF. To understand the usefulness of these mutation operators in \our, we develop a variant, dubbed \texttt{CFG-}\our, that uses a random sampling method based on the CFG w.r.t. the underlying injection attack to replace the mutation operators in \our. In particular, we investigate \texttt{CFG-}\our\ for all three surrogate classifiers considered in this paper.

\subsubsection{Results}
\label{sec:results_rq2}

\pref{tab:rq3} shows the performance of \texttt{CFG-}\our\ under different WAFs by using three surrogate classifiers. As we expected, the six mutation operators play a major role for generating test cases. It is clear to see that the performance of \our, without using the mutation operators, is degraded in all scenarios. In particular, this variant even generates fewer valid test inputs than the \texttt{STEA} methods in some cases, e.g., on SQLi, PHPi and XMLi. As shown in~\pref{tab:rq3}, \texttt{CFG-}\our\ suffers more than 50\% performance loss on SQLi and XMLi in most of the scenarios. This indicates that the corresponding WAFs are more vulnerable than those mutated SQLi and XMLi test inputs. On the contrary, \texttt{CFG-}\our\ generally outperforms the corresponding \texttt{STEA} methods on XSSi and HTMLi, which are featured in a longer length of the injection string. This indicates that the bidirectional translation can lead to more diversified test inputs so as to make up the missing mutation operators to a certain extent. 

\vspace{1em}
\noindent
\framebox{\parbox{\dimexpr\linewidth-2\fboxsep-2\fboxrule}{
    \textbf{\underline{Response to RQ2:}} \textit{From the empirical results discussed in this subsection, we confirm the importance of our proposed six mutation operators. They can bring more diversity into the \texttt{MTEA} thus leading to an increased number of valid test inputs.
    }}}

\subsection{Investigation of the Impacts of the Surrogate Classifier in \our}
\label{sec:rq3}

\subsubsection{Methods}
\label{sec:methods_rq3}

According to the experiments in~\pref{sec:results_rq2}, we have already validated the effectiveness of those mutation operators for offspring reproduction in the \texttt{MTEA} of \our. How to select the elite solutions to either survive to the next generation or to construct the mating pool for offspring reproduction is another important component of an evolutionary algorithm. In \our, the selection process is guided by the surrogate classifier that predicts the chance of a candidate test input for bypassing the underlying WAF. To address \textbf{RQ3}, we replace this mechanism with a random selection, dubbed \texttt{RAN}. More specifically, it randomly picks up the mating parents from the mating pool, so as the survival of parents and offspring. Note that we still apply the bidirectional translation between different types of injection attack within the mating pool.

\subsubsection{Results}
\label{sec:results_rq3}

From the comparison results shown in~\pref{tab:rq3}, it is clear to see that the performance of \texttt{RAN} is outperformed by our proposed \our\ in all scenarios. In addition, \texttt{STEA} can even find more valid test inputs than \texttt{RAN} in some cases such as the XMLi on \texttt{ModSecurity} and \texttt{Lua-Resty-WAF}. This observation confirms the importance of elitism in an evolutionary algorithm. In other words, without the guidance of an appropriate fitness function (i.e., the surrogate classifier in \our), the evolutionary search suffers from a lack of selection pressure thus is less effective. Another interesting observation from our experiments is that the capability of generating valid test inputs is partially related to the length of the characters w.r.t. the corresponding injection. More specifically, if an injection string is long such as XSSi and HTMLi, a sufficient selection pressure becomes important to guide the evolutionary search. This explains the better performance achieved by \texttt{STEA} w.r.t. \texttt{RAN} on XSSi and HTMLi. On the other hand, if the length of an injection string is short such as OSi, the translation between different types of injection attack can be highly beneficial to the generation of valid test inputs.

\vspace{1em}
\noindent
\framebox{\parbox{\dimexpr\linewidth-2\fboxsep-2\fboxrule}{
    \textbf{\underline{Response to RQ3:}} \textit{From the comparison results discussed in this subsection, we confirm the importance of the surrogate classifier as an alternative of the fitness function. It provides a sufficient selection pressure to guide the evolutionary search process. This is critical for problems with a large search space.
}}}

\subsection{An Qualitative Study of the Translation in \our}
\label{sec:rq4}

\subsubsection{Methods}
\label{sec:methods_rq4}

From the discussion in~\pref{sec:rq1}, we have confirmed the effectiveness and outstanding performance of \our\ for generating valid test inputs for different types of injection attacks. \textbf{RQ3} aims to step further and to understand whether the semantic meanings between different types of injection attack have indeed been learned and exploited. In this section, we plan to qualitatively analyze some examples selected from our experiments, as well as some that go beyond our expectation.

\subsubsection{Results}
\label{sec:results_rq4}

By investigating the intermediately translated and evolved test inputs together with their parent inputs in \our, we found many examples similar to those listed in~\pref{tab:exp}. This confirms that creating tautology is the most common way in injection attack. Nevertheless, we have also identified some \lq unusual\rq\ translations and evolution. In the following paragraphs, we will discuss some selected examples.

\begin{code-example}
\textbf{\underline{\textit{Example 1}:}} from SQLi 
(\textcolor{blue}{\texttt{\rq /**/or/**/\lq 1\rq=\lq 1\rq --}}) to XSS (\textcolor{blue}{\texttt{\textgreater{}\textless{}s\%43\%72\%49pt\textgreater{}\%61\%6c\%65\%72\%74\%28}}

\textcolor{blue}{\texttt{\%31\%29\textless{}/s\%43ri\%70\%74\textgreater{}\textless{}!--}}).
\end{code-example}

In the \textbf{\textit{Example 1}}, the parent input for SQLi is translated and evolved to another test input for XSSi equivalent to \textcolor{blue}{\texttt{\textgreater{}\textless{}script\textgreater{}alert(1)\textless{}/script\textgreater{}\textless{}!--}}, some of the characters are in ASCII encoding. We notice that \texttt{\textsc{DaNuoYi}} did not pick up the semantic meaning of tautology attack (as in the SQLi), but it does take the other piece of semantic information from the SQLi. In particular, the previous statement is closed by \textcolor{blue}{\texttt{'}} in SQLi, so that \textcolor{blue}{\texttt{or '1'='1'}} can be injected successfully. The same operation is used in XSSi which using \textcolor{blue}{\texttt{>}} to close the previous statement.

\begin{code-example}
\textbf{\underline{\textit{Example 2}:}} from OSi 
(\textcolor{blue}{\texttt{0 ; sleep \%20 1 |}}) to SQLi (\textcolor{blue}{\texttt{)/**/;select sleep (2) \#}}).
\end{code-example}

In the \textbf{\textit{Example 2}}, \texttt{\textsc{DaNuoYi}} learned that the function \textcolor{blue}{\texttt{sleep \%20 1}}\footnote{\textcolor{blue}{\texttt{\%20}} is a \textit{blank space} in ASCII encoding.} in OSi is similar to the function \textcolor{blue}{\texttt{sleep (2)}} in SQLi. In particular, it conducted two things when translating and evolving such test input from SS to SQLi. First, it changes the separator from \textcolor{blue}{\texttt{|}} to \textcolor{blue}{\texttt{\#}}. Then, it takes a different syntax into account, i.e., the sleep function of SQLi is \textcolor{blue}{\texttt{sleep()}} but not for the OSi. This is a typical example in which some semantic knowledge is captured by \our\ while being aware of their syntactical differences.

\begin{code-example}
\textbf{\underline{\textit{Example 3}:}} from SQLi 
(\textcolor{blue}{\texttt{\lq)/**/\&\& not/**/false }} 

\textcolor{blue}{\texttt{or(\rq}}) to PHPi (\textcolor{blue}{\texttt{O:7:``Example”:1:\{s:3:``var”;}}

\textcolor{blue}{\texttt{s:10:``phpinfo();”;\}}}).
\end{code-example}

The \textbf{\textit{Example 3}} is one of the most surprising examples we found from the results of \our. At the first glance, the tautology in SQLi (due to the \textcolor{blue}{\texttt{not false}} clause) may seem to be irrelevant to the serialized code injection in PHPi. Yet, if we inspect closely, it is clear that the PHPi aims to access some information through the \textcolor{blue}{\texttt{phpinfo()}} call somewhere in the \texttt{eval} function. This matches with the ultimate purpose of creating a tautology, i.e., accessing or revealing some unauthorized information by creating something that is always true. This is beyond our expectation, as it suggests that \our\ is not only able to interpret the literal meaning in the test inputs, but also some of the semantics related to the ultimate goal of an attack.

\vspace{1em}
\noindent
\framebox{\parbox{\dimexpr\linewidth-2\fboxsep-2\fboxrule}{
        \textbf{\underline{Response to RQ4:}} \textit{From the discussion in this subsection, we appreciate the ability of \our\ to share and exploit promising test inputs from different types of injection attacks. By doing so, it can discover more hidden vulnerabilities of the WAF as we showed in Sections~\ref{sec:rq1} and~\ref{sec:rq2}.}
    }}

\subsection{Investigation of the Impacts of CFG in \our}
\label{sec:rq5}

\subsubsection{Methods}
\label{sec:methods_rq5}

As discussed in~\pref{sec:grammar}, the CFG used in the data gathering and profiling step is derived from some existing injection attack examples. According the corresponding CFG for an injection type, the rule-based injection search (RIS) method introduced in~\pref{sec:grammar} plays as the source to seed some initial test inputs for training the surrogate classifier. In this case, a natural question is whether this RIS method guided by the CFG adequate to serve the purpose of test input generation? To address \textbf{RQ5}, we use the CFG as the template to generate $100,000$ test inputs for each injection type. In particular, we are mainly interested in the statistics of the number of non-duplicated test inputs and those bypassing the underlying WAFs.

\begin{table*}[htbp]
  \centering
  \caption{The validity statistics of injection cases generated by rule-based injection search according to a given CFG.}
      \resizebox{\linewidth}{!}{
       \begin{tabular}{c|c|c|c|c}
    \hline
    \multirow{1}[4]{*}{Injection type} & \multirow{1}[4]{*}{\# of non-duplicated test inputs (percentage)} & \multicolumn{3}{c}{\# of bypassed test inputs (success rate)} \\
\cline{3-5}     &       & \texttt{ModSecurity} & \texttt{Ngx-Lua-WAF} & \texttt{Lua-Resty-WAF} \\
    \hline
    SQLi   & 25079 (25.1\%) & 2082  (8.3\%)& 24287 (96.8\%) & 5502 (21.9\%) \\
    \hline
    OSi    & 6534 (6.5\%)& 2826 (43.3\%) & 6534 (100\%) & 5766 (88.2\%) \\
    \hline
    PHPi   & 48093 (48.1\%)&37593 (78.2\%) & 20366 (89.0\%) & 22880 (47.6\%) \\
    \hline
    XMLi   & 25042 (25.0\%)& 2110 (8.4\%) & 24250 (96.8\%) & 4997 (20.1\%) \\
    \hline
    XSSi    & 98931 (98.9\%)& 16978 (17.2\%) & 23419 (23.7\%) & 35824 (36.2\%) \\
    \hline
    HTMLi  & 99535 (99.5\%)& 18371 (18.6\%) & 55439 (55.7\%) & 19510 (19.6\%) \\
    \hline
    \end{tabular}
    }
  \label{tab:cfg_result}%
\end{table*}%

\subsubsection{Results}
\label{sec:results_rq5}

According to the statistical results shown in~\pref{tab:cfg_result}, we can see that the percentage of the non-duplicated test inputs is less than $50\%$ for SQLi, XMLi, OSi, and PHPi. Especially for OSi, the corresponding percentage is merely around $6.5\%$. However, it is surprising to see that the success rates to bypass all three WAFs for OSi are very promising. This can be explained as the relatively small search space of OSi as discussed in~\pref{sec:rq1} that renders the test inputs generated by the corresponding CFG duplicated. In contrast, the success rates of bypassing the WAFs for SQLi, XMLi and PHPi vary significantly case by case. For example, both SQLi and PHPi experience a hard time in \texttt{ModSecurity} of which the success rates of the bypassed test inputs generated by the corresponding CFGs are less than $10\%$; whereas the success rates for \texttt{Ngx-Lua-WAF} are always high. On the other hand, we can see that the percentage of non-duplicated test inputs generated by the RIS methods for XSSi and HTMLi is extremely high with $98.9\%$ and $99.5\%$, respectively. This can be explained as the fact that both HTMLi and XSSi tend to be long strings leading to a large search space, i.e., an exponentially increased number of potential combinations of different strings. Partially due to this reason, the success rates to bypass all three WAFs for those generated test inputs are unfortunately low. This suggests that the RIS can hardly find effective injection cases in a large search space.

In addition, we compare the RIS method with \our\ under the same amount of computational search budget for generating test inputs. Since the RIS method cannot work in a multi-task manner, we run a test input generation routine for one injection type at a time, as done in \texttt{STEA}. From the comparison results shown in~\pref{tab:rq3}, we find that the performance of the RIS method is significantly worse than that of \our\ in all cases, especially for \texttt{ModSecurity} and \texttt{Lua-Resty-WAF}, the relatively more challenging WAFs. In particular, it is worth noting that the RIS method is comparable with \texttt{STEA} in many cases. This also supports the importance of our multi-task mechanism that complements each other when generating different types of injection attacks. Let us look at the evolutionary trajectories shown in~\pref{fig:traj_mod} to~\pref{fig:traj_ngx}, we can see that \our\ keeps on generating more effective test inputs after being seeded by the RIS as the initial test inputs. This observation confirms the importance and usefulness of the follow-up evolution and translation in \our.

\vspace{1em}
\noindent
\framebox{\parbox{\dimexpr\linewidth-2\fboxsep-2\fboxrule}{
        \textbf{\underline{Response to RQ5:}} \textit{There are two takeways from the experiments in this subsection. First, the CFG itself can be used to generate test inputs for a given injection type. However, its effectiveness is hardly guaranteed especially when handling a large search space. Second, by leveraging the semantic information associated with each injection attack along with the multi-task paradigm, \our\ can develop more effective test inputs for different types of injection attacks simultaneously.}
    }}

\section{Threats to Validity}
\label{sec:tov}

Threats to construct validity may raised from the way we measure the effectiveness of \texttt{\textsc{DaNuoYi}}. In this work, we use the number of generated test inputs that bypass the WAF as the main metric, following the suggestion from prior work~\cite{AppeltNPB18}. To improve reliability of the conclusions, we have also applied Wilcoxon rank-sum test to evaluate the statistical significance.

The parameter settings in the experiments, e.g., the dimension of the \texttt{Word2vec} and the number of generations in MTEA, may cause threats to internal validity. Indeed, we acknowledge that a different set of settings may lead to other results. However, we set these parameter values based on the results of both automatic and manual hyper-parameter tuning.


To ensure external validity, we evaluate \texttt{\textsc{DaNuoYi}} on three widely used real-world WAFs and six most prevalent types of injection attack. We have also run \texttt{\textsc{DaNuoYi}} under three alternative classifiers, i.e., LSTM, RNN, and GRU. To mitigate evaluation bias, we repeat each experiment run $10$ times. Nonetheless, we do agree that additional subject WAFs are useful for future work.


\section{Related Works}
\label{sec:related}

Over the past decade, several white-box, static and model-based approaches have been proposed on testing injection vulnerability based on specific syntax, mainly target exclusively for SQLi. Among others, Halfond and Orso~\cite{HalfondO05} propose AMNESIA, which generates test input for SQLi based on static code analysis. Similarly, Mao et al.~\cite{MaoG16} also seek to generate SQLi test inputs based on a pre-defined finite automata. However, these approaches are restricted by the given rules and require full access to the source code, which is what limits their testing ability as already shown in prior work~\cite{AppeltNPB18,JanPAB19,LiuLC20}.

In the industry, \texttt{SQLMap} is a well-known rule-based testing tool for SQLi, which is also applicable on WAF. However, it is fundamentally different from \texttt{\textsc{DaNuoYi}} and difficult to be experimentally compared in a fair manner, because: 

\begin{itemize}
    \item \texttt{SQLMap} focuses on SQLi only. \texttt{\textsc{DaNuoYi}}, in contrast, works on any type of injection attack.
    
    \item \texttt{SQLMap} stops as soon as it find a test input that can bypass the WAF while \texttt{\textsc{DaNuoYi}} would continue to evolve till the pre-defined resources are exhausted.
    
    \item Prior work (e.g., Liu et al.~\cite{LiuLC20}) has shown that \texttt{SQLMap} is significantly inferior to single-task learning and search-based testing methods, which are the peer methods that we have quantitatively compared for \textbf{RQ1} and \textbf{RQ2}.
\end{itemize}

Other black-box, learning and search-based fuzzers for injection testing also exist. For example, to discover SQLi vulnerabilities on WAF, Demetrio et al.~\cite{DemetrioVCL20} combine a random search fuzzer with strongly-typed syntax for testing. Similarly, Appelt et al.~\cite{AppeltPB17,AppeltNPB18} leverage evolutionary algorithm supported by a learnt surrogate model to generate test inputs. Liu et al.~\cite{LiuLC20} propose \texttt{DeepSQLi}, a method that uses techniques from NLP to learn and test SQLi vulnerabilities in web applications, including the WAFs. Edalat et al. \cite{EdalatSG18} develops an Android-oriented SQL injection detection tool to protect Android applications, while this tool aims at function-level attack instead of multiple types of injection attack. Uwagbole et al. \cite{UwagboleBF17} propose to generate SQL injections by leveraging attack patterns derived from existing SQL injections. However, this method relies on the reserved keywords of SQL and is hard to adapt to other injection types. Eassa et al. \cite{EassaEES18} argues that the importance of defending injection attack on NoSQL is underestimated and try to detect injections by developing an independent module based on PHP. They share some of the same motivations as \texttt{\textsc{DaNuoYi}} such that every test input has its own semantic information. For other injection types, Jan et al.~\cite{JanPAB19} exploit an evolutionary algorithm, empowered by a specifically designed distance function, to generate test inputs for XMLi on WAF. Evolutionary fuzzing for testing XSS has also been explored~\cite{DucheneGRR12}. 

On the other hand, many recent works focus on injection detection (e.g., code injection attack and SQL injection detection). Zuech et al. \cite{ZuechHK21} argues that using ensemble classifiers to predict potential SQL injections. However, the test injections used in work are prepared as a dataset that is not designed to find new injections. Thang \cite{Thang20} also adopts three existing dataset to predict the malicious HTTP requests. Jahanshahi et al. \cite{JahanshahiDE20} presents SQLBlock, a plugin for PHP \& MySQL-based Web applications, to prevent SQL injection without any modification on existing Web applications. And the test case preparation method is not reusable for other injection types. 

As mentioned, the above work neither publishes the source code (or the code contains severe errors) nor has readily available tools for directly quantitative comparisons. 
However, given the common techniques that underpin those tools, they are the resemblances of the single-task counterparts we evaluated for \textbf{RQ1}. It is clear that \texttt{\textsc{DaNuoYi}} differs from all the above work in the sense that:

\begin{itemize}
     \item It works on any type of injection attack as opposed to one.
    \item It learns, translates and exploits the common semantic information across different types of injection attack. This, as we have shown in Section~\ref{sec:evaluation}, allows \texttt{\textsc{DaNuoYi}} to find more bypassing test inputs on the WAF.
\end{itemize}




\section{Conclusion}
\label{sec:conclusion}

In this paper, we propose \texttt{\textsc{DaNuoYi}}, a multi-task end-to-end fuzzing tool that simultaneously generates test inputs for any type of injection attacks on WAF. \texttt{\textsc{DaNuoYi}} trains a classifier to predict the likelihood of a test input bypassing the WAF, and a pair of translation models between any two types of injection attack. These then equip the proposed multi-task evolutionary algorithm, which is the key component that realizes the multi-tasking in \texttt{\textsc{DaNuoYi}}, with the ability to share the most promising test inputs for different injection types. Through experimenting on three real-world open-source WAFs, three classifiers, and six types of injection attack, we show that, in \texttt{\textsc{DaNuoYi}}, both the multi-task translation and multi-task search are effective in handling and transferring the common semantic information for different injection types, allowing it to produce up to $3.8\times$ more bypassing test inputs than its single-task counterparts.

Future opportunities from this work are fruitful, including extending \texttt{\textsc{DaNuoYi}} beyond injection testing and a better synergy with the concept from reinforcement leaning. Since different reproduction operators can have distinct behaviors, it will be interesting to develop an adaptive mechanism to choose the most appropriate operator for offspring reproduction~\cite{CaoKWL12,CaoKWL14,LiKWTM13,LiK14,LiFK11,LiFKZ14,LiKWCR12,LiWKC13,CaoKWLLK15,LiDZ15,LiXT19}. In addition, we can consider more complicated scenarios with multiple non-functional objectives~\cite{LiZZL09,LiZLZL09,LiKZD15,WuLKZZ17,LiDZZ17,LiKD15,LiDZK15,LiLLM21,CaoWKL11,WuLKZ20,WuKJLZ17,WangYLK21,LiKCLZS12,ChenLY18,ZouJYZZL19,BillingsleyLMMG19,LiZ19,ShanL21,YangHL21,BillingsleyLMMG21,LiC22} and challenging constraints~\cite{LiCFY19}. In addition, we can also take the decision makers' preference information~\cite{Li19,LiCSY19,WuLKZZ19,LiDY18,WuKZLWL15,LiDAY17,LiLDMY20,LaiL021,XuLA22} into the evolutionary search to generate more personalized examples~\cite{GaoNL19}.


\section*{Acknowledgment}
This work was supported by UKRI Future Leaders Fellowship (MR/S017062/1), EPSRC (2404317), NSFC (62076056), Royal Society (IES/R2/212077) and Amazon Research Award.

\bibliographystyle{IEEEtran}
\bibliography{IEEEabrv,mybib}

\end{document}